\documentclass[sigconf]{acmart}

\AtBeginDocument{%
  \providecommand\BibTeX{{%
    \normalfont B\kern-0.5em{\scshape i\kern-0.25em b}\kern-0.8em\TeX}}}

\copyrightyear{2022}
\acmYear{2022}
\setcopyright{acmcopyright}
\acmConference[CCS '22]{Proceedings of the 2022 ACM SIGSAC Conference on Computer and Communications Security}{November 7--11, 2022}{Los Angeles, CA, USA}
\acmBooktitle{Proceedings of the 2022 ACM SIGSAC Conference on Computer and Communications Security (CCS '22), November 7--11, 2022, Los Angeles, CA, USA} \acmPrice{15.00}
\acmDOI{10.1145/3548606.3560566}
\acmISBN{978-1-4503-9450-5/22/11}

\usepackage{algorithm}
\usepackage{algorithmic}
\usepackage{mathrsfs}
\usepackage{marvosym}
\usepackage{multirow}
\usepackage{graphicx}
\usepackage{subfigure}
\usepackage{adjustbox}
\usepackage{colortbl}  
\usepackage{xcolor}
\def\eg{\emph{e.g.}}
\def\ie{\emph{i.e.}}
\def\etc{\emph{etc}}

\begin{document}

\title{Harnessing Perceptual Adversarial Patches for Crowd Counting}

\author{Shunchang Liu}
\authornotemark[1]
\affiliation{%
  \institution{Beihang University}
  \city{Beijing}
  \country{China}}
\email{liusc@buaa.edu.cn}

\author{Jiakai Wang}
\authornote{indicates equal contribution.}
\affiliation{%
  \institution{Zhongguancun Laboratory}
  \city{Beijing}
  \country{China}}
\email{wangjk@mail.zgclab.edu.cn}

\author{Aishan Liu}
\authornote{indicates corresponding author.}
\affiliation{%
  \institution{Beihang University}
  \city{Beijing}
  \country{China}}
\email{liuaishan@buaa.edu.cn}

\author{Yingwei Li}
\affiliation{%
  \institution{Johns Hopkins University}
  \city{Baltimore}
  \country{United States}}
\email{yingwei.li@jhu.edun}

\author{Yijie Gao}
\affiliation{%
  \institution{Beihang University}
  \city{Beijing}
  \country{China}}
\email{yijie422@buaa.edu.cn}

\author{Xianglong Liu}
\affiliation{%
  \institution{Beihang University}
  \city{Beijing}
  \country{China}}
\email{xlliu@buaa.edu.cn}

\author{Dacheng Tao}
\affiliation{%
  \institution{JD Explore Academy}
  \city{Beijing}
  \country{China}}
 \affiliation{%
  \institution{The University of Sydney}
  \city{Sydney}
  \country{Australia}}
\email{dacheng.tao@gmail.com}

\begin{abstract}
Crowd counting, which has been widely adopted for estimating the number of people in safety-critical scenes, is shown to be vulnerable to adversarial examples in the physical world (\eg, adversarial patches). Though harmful, adversarial examples are also valuable for evaluating and better understanding model robustness. However, existing adversarial example generation methods for crowd counting lack strong transferability among different black-box models, which limits their practicability for real-world systems. Motivated by the fact that attacking transferability is positively correlated to the model-invariant characteristics, this paper proposes the \emph{Perceptual Adversarial Patch (PAP)} generation framework to tailor the adversarial perturbations for crowd counting scenes using the model-shared perceptual features. Specifically, we handcraft an adaptive crowd density weighting approach to capture the invariant scale perception features across various models and utilize the density guided attention to capture the model-shared position perception. Both of them are demonstrated to improve the attacking transferability of our adversarial patches. Extensive experiments show that our PAP could achieve state-of-the-art attacking performance in both the digital and physical world, and outperform previous proposals by large margins (at most {+685.7 MAE} and {+699.5 MSE}). Besides, we empirically demonstrate that adversarial training with our PAP can benefit the performance of vanilla models in alleviating several practical challenges in crowd counting scenarios, including generalization across datasets (up to {-376.0 MAE} and {-354.9 MSE}) and robustness towards complex backgrounds (up to {-10.3 MAE} and {-16.4 MSE}) \footnote{Our code can be found in \url{https://github.com/shunchang-liu/PAP-Pytorch}.}.

\end{abstract}



\maketitle
\section{Introduction}

\begin{figure}[!t]
\centering

\includegraphics[width=1.0\linewidth]{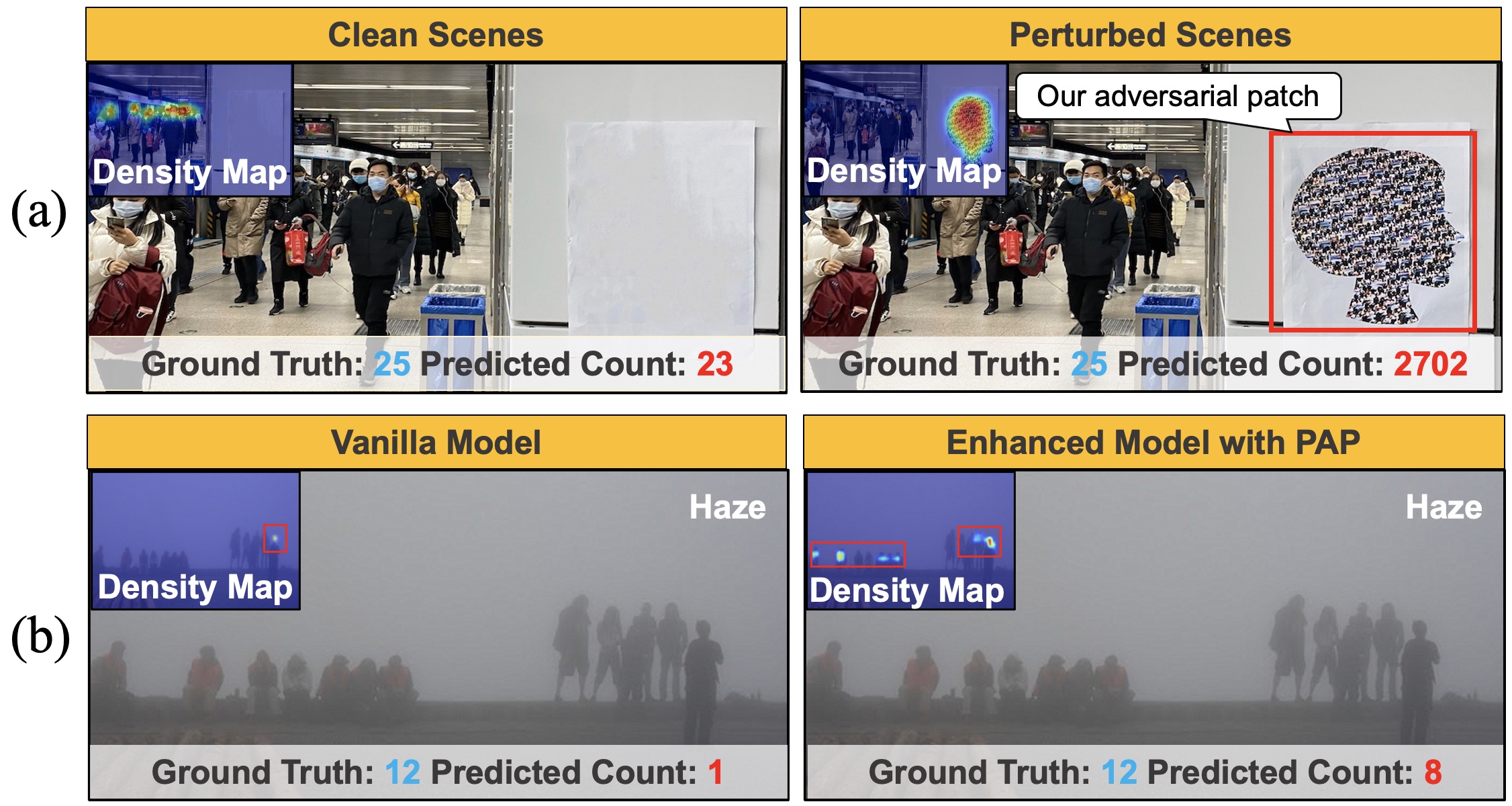}
\caption{Bidirectional role of our adversarial patches in the crowd counting scenario. (a) Physical world attacks (left: clean scene; right: perturbed scene). Our patch can lead the crowd counting system to seriously wrong predictions. (b) Performance improvement with PAP (left: vanilla model; right: enhanced model trained with PAP). Model robustness towards complex backgrounds (\eg, Haze) is improved by adversarial training with our patches.} 
\label{fig:app-intro}
\vskip -4ex
\end{figure}

Crowd counting, which estimates the number of people in unconstrained scenes, is becoming increasingly important in many safety-critical scenarios in practice (\eg, pedestrian density monitoring). Up to now, research has focused on designing different crowd counting methods, including detection-based approaches \cite{lin2001estimation,li2008estimating,zhao2003bayesian}, count regression approaches \cite{chan2009bayesian,chen2012feature,wang2015deep,chen2013cumulative}, and density-map-estimation-based methods \cite{zhang2016single,onoro2016towards,boominathan2016crowdnet,sheng2016crowd,li2018csrnet,liu2019context,zhang2019relational,ma2019bayesian,wang2020distribution,song2021choose}. The last has become the de facto solution for the crowd counting task due to its insensitivity to occlusion and stability for large crowd scenes. In general, given an input image, this type of approach first generates a 2D crowd density map and then subsequently estimates the total number of the crowd by summing the density values across all spatial locations of the density map. 

With increasing deployment of intelligent crowd counting devices in the safety-critical scenarios, their vulnerability has attracted considerable attention and become a growing concern for the public. Unfortunately, current estimation-based crowd counting models are highly vulnerable to adversarial examples, \ie, small perturbations that are imperceptible to humans but can easily lead deep neural networks to make wrong predictions \cite{szegedy2013intriguing}. Though harmful, adversarial attacks could also be used to evaluate model robustness and provide valuable insights into the blind spots of deep learning models. In contrast with $L_p-$norm based attacks \cite{szegedy2013intriguing, goodfellow6572explaining}, adversarial patch \cite{brown2017adversarial}, a type of perturbation confined into a small patch region without an $\epsilon-$ball constraint, has a much higher application value in the real world due to its strong resistance to physical influences. However, as the only adversarial patch study for crowd counting models, \cite{wu2021towards} performs weak transferable attacks, which limits its ability to evaluate the robustness of black-box crowd counting systems in practice.

\begin{figure*}[!ht]
\begin{center}

\includegraphics[width=1.0\linewidth]{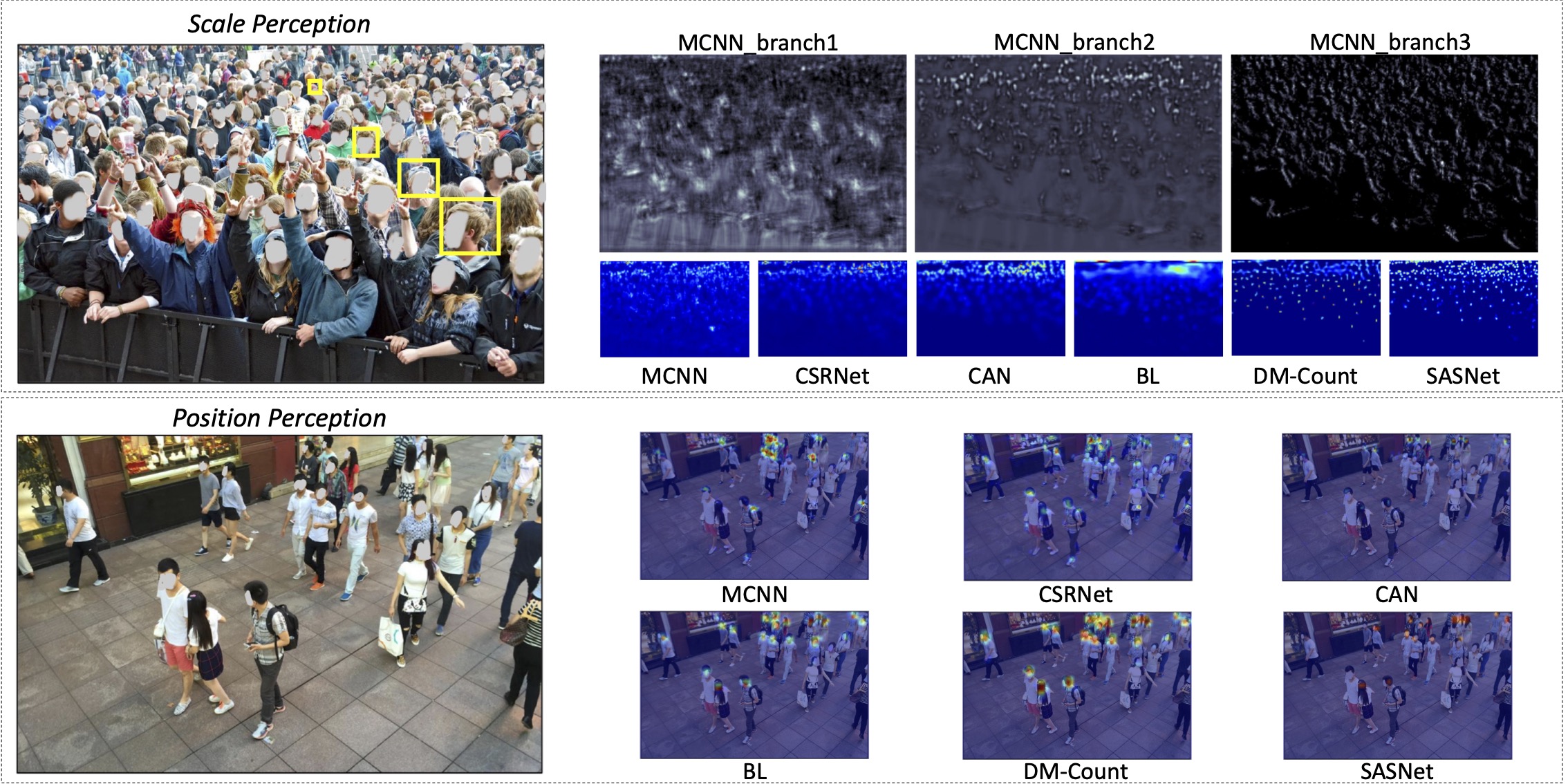}
\end{center} 
\caption{Perceptual properties of density-map-estimation-based crowd counting models. For scale perception, we show the feature maps in three branches of MCNN and the density maps of six models. Branches with different receptive fields capture different scale features, and thus various model structures lead to multiple scale preferences in the density map. For position perception, we show the model attention maps through Grad-CAM \cite{selvaraju2017grad}. All models have similar spatially dispersed attention patterns for the same crowd position.} 
\label{fig:2}
\vskip -2ex
\end{figure*}

Recent studies \cite{dong2019evading,lennon2021patch} have shown that model-invariant characteristics greatly influence transferable attacks in vision tasks. In light of this, we aim to find those intrinsic characteristics that are shared between models for generating adversarial patches with strong transferability. For crowd counting, we reached \textbf{two key insights}: (1) Different models tend to show different perceptual preferences for different crowd scales, \ie, multiple scale perceptions. Due to the different structures, various models with different receptive fields can hardly capture consistent scale representations. Therefore, a model trained with specific object scales (\eg, the sizes of human heads) always performs well on its preferred scale while it is difficult to make correct predictions on others. (2) Different models show similar attention patterns at the same crowd positions, \ie, shared position perception. Almost all density-estimation-based models rely on head features for crowd prediction. In other words, they have similar attention patterns to the position of human heads. Figure \ref{fig:2} illustrates the above observations.

Thus, based on the above investigations, we propose the \emph{Perceptual Adversarial Patch (PAP)} generation framework to learn model-invariant features by exploiting the model scale and position perceptions, thus promoting the transferability of our adversarial patches. (1) As for \textbf{scale perception}, PAP introduces the adaptive density during training to dynamically adjust the contribution of features with different scales, which helps to capture the scale invariance between models. In particular, we automatically enhance the contribution of the crowd scale features that are not captured well by the specific target model, so that the adversarial patches can be optimized with the complete scale features, \ie, the adversarial patches could adapt to models with different crowd scale perceptions. (2) Regarding \textbf{position perception}, PAP draws the model-shared attention of the target model from the spatially dispersed crowd patterns to the patch region, which helps to capture position invariance among models. Specifically, we utilize density-based gradients to obtain the attention map and strengthen the salient degree of the patch region. Thus, we could force the position perception of different models to focus on the patch. Overall, as shown in Figure \ref{fig:app-intro} (a), our approach can generate strongly transferable adversarial patches in the physical world.

Furthermore, while most studies have found that adversarial training will reduce the model's performance on the original task \cite{madry2017towards,tsipras2018robustness}, we found an intriguing effect that \textbf{benefits model performance} for crowd counting by training with our adversarial patches. Since the generated adversarial patches consist of model-invariant characteristics (\ie, scale perception and position perception), adversarial training with our patches can force the vanilla model to better focus on crowds at perception level. We empirically demonstrate that it could benefit the vanilla models for better generalization across datasets and better robustness towards complex backgrounds (as shown in Figure \ref{fig:app-intro} (b)).

To sum up, our \textbf{contributions} are as follows:
 \begin{itemize}
     \item We proposed a Perceptual Adversarial Patch (PAP) generation framework that exploits the inherent perceptual properties to capture the model-invariant features for attacking the real-world crowd counting systems.
     \item We designed the adaptive density and guided attention to capture scale and position perceptions, which could improve the transferable attacking ability of the adversarial patches across multiple crowd counting models in various structures.
     \item Besides, we empirically demonstrated that our generated adversarial patches could be utilized for promoting the vanilla model's robustness in several aspects (\eg, generalization across datasets and robustness towards complex backgrounds) via the adversarial training scheme.
     \item Extensive experiments in the digital and physical world demonstrated that our PAP achieves the state-of-the-art attacking ability and outperforms other baselines by large margins (at most \textbf{+685.7 MAE} and \textbf{+699.5 MSE}). In addition, adversarial training with our PAP can improve the model performance by at most \textbf{-376.0 MAE}, \textbf{-354.9 MSE} for generalization across datasets, and \textbf{-10.3 MAE}, \textbf{-16.4 MSE} for robustness towards complex backgrounds.
 \end{itemize}

\section{Related Works and Preliminary}
\label{sec:background}
\subsection{Crowd Counting}
Image or video-based crowd counting aims to automatically estimate the number of people in unconstrained scenes. Early work mainly focuses on detection-based methods \cite{lin2001estimation,li2008estimating,zhao2003bayesian} and count regression methods \cite{chan2009bayesian,chen2012feature,wang2015deep,chen2013cumulative}, which either show unsatisfactory results in extremely dense crowds or have weak interpretability due to ignoring the key information in the dot annotation maps. Nowadays, density-map-estimation-based approaches \cite{zhang2016single,onoro2016towards,boominathan2016crowdnet,sheng2016crowd,li2018csrnet,liu2019context,zhang2019relational,ma2019bayesian,wang2020distribution,song2021choose}, which we focus on, have been widely used due to their better performance. They take images or videos as inputs and predict the crowd density maps to estimate the number of people. Formally, given an input image $\mathbf{x}$, a model $f_\Theta$ is designed to approximate the ground truth density map $\mathbf I$ by solving the following optimization problem:
\begin{equation}
    \arg\min_\Theta{\frac{1}{2N}\sum_{i=1}^{N}{||f_{\Theta}(\mathbf x_i)-\mathbf I_i||^2_2}},
\end{equation}
where $N$ denotes the number of input samples.

Several methods have been proposed as solutions to this problem. For instance, \cite{zhang2016single} designed a Multi-column Convolutional Neural Network (MCNN) utilizing three network branches with different kernel sizes to map the image to its density map. \cite{li2018csrnet} replaced the multi-branch structure with dilated convolution and proposed an end-to-end Congested Scene Recognition Network (CSRNet). Further, \cite{liu2019context} introduced an adaptively Context-Aware Network (CAN) to capture the contextual information. From an optimization view,  \cite{ma2019bayesian} designed a Bayesian Loss (BL) to construct a density contribution probability model and \cite{wang2020distribution} proposed optimal transport loss and total variation loss for Distribution Matching (DM-Count). Recently, \cite{song2021choose} proposed a Scale-Adaptive Selection Network (SASNet) which can automatically learn the internal correspondence between the scales and the feature levels and further proposed a pyramid region awareness loss to fix the most hard sub-regions. These estimation-based methods can be roughly divided into two categories based on the branches used for feature extraction: multi-column strategies, \eg, \cite{zhang2016single,onoro2016towards,boominathan2016crowdnet,zhang2019relational,song2021choose}, and single-column strategies, \eg, \cite{sheng2016crowd,li2018csrnet,liu2019context,ma2019bayesian, wang2020distribution}.

Though having achieved promising results, \cite{gao2020cnn} pointed out that current crowd counting models still suffer from multiple challenges. Specifically, current methods show \textbf{unsatisfactory generalization across datasets}. The overfitting problem has always been the hot potato in the field of deep learning. As for the estimation-based crowd counting methods, the performance of the predictor will inevitably reduce when generalizing the model trained on the specific data distribution to unseen scenes with non-uniform distributions. The reduction will cause sub-optimal results, though it may not lead to a collapse of the model. Moreover, they have \textbf{weak robustness for complex backgrounds}. That is, it is easily influenced by natural noises, \eg, adverse weather (rain, snow, haze, \etc.), and objects with similar densities, \eg, hard samples that are similar to crowds (leaves, birds, \etc). These drawbacks inject potential risks into real-world crowd counting systems.

\subsection{Adversarial Attacks}
Adversarial examples are inputs intentionally designed to mislead DNNs but are imperceptible to humans \cite{szegedy2013intriguing,goodfellow6572explaining}. A long line of work has been devoted to performing adversarial attacks in different scenarios by generating imperceptible perturbations \cite{goodfellow6572explaining,madry2017towards,chen2017zoo,brendel2017decision,ilyas2018black,dong2018boosting,athalye2018obfuscated,dong2019efficient,dong2019evading,xie2019improving,croce2020autoattack}. These adversarial attacking methods are mainly divided into white-box and black-box manners. For \textbf{white-box attacks}, adversaries have complete knowledge of the target model and can fully access it. For example, \cite{szegedy2013intriguing} first introduced the L-BFGS method to generate adversarial examples. Subsequently, \cite{goodfellow6572explaining} proposed the Fast Gradient Sign Method (FGSM), and \cite{madry2017towards} improved it and proposed the Projected Gradient Decent (PGD) method, which is currently the strongest first-order attack. All of them depend on access to the gradients of target models. For \textbf{black-box attacks}, adversaries have limited model knowledge and can not directly access the model. Black-box attacks can be divided into three categories, \ie, score-based, decision-based, and transfer-based. The score-based \cite{chen2017zoo,ilyas2018black} and decision-based \cite{brendel2017decision,dong2019efficient} attacks rely on querying either the output scores or labels of the target network, which limits their usability in the physical world. The transfer-based attacks generate adversarial perturbations on a source model and then transfer them to the unknown target model. A series of approaches \cite{dong2018boosting,dong2019evading,xie2019improving} have been proposed to improve the attack transferability among different models and achieve substantial results in the digital world. However, their attacking abilities will degenerate significantly when introduced into the physical world.

Besides perturbations, adversarial patches \cite{brown2017adversarial}, where noises are confined to a small and localized patch region, have emerged for its easy accessibility in real-world scenarios. They have been widely studied and applied to attack different real-world applications. \cite{eykholt2018robust} mixed the attacking noises into the black and white stickers to attack the stop sign recognition devices. \cite{Liu2019Perceptual} proposed the PS-GAN framework to generate scrawl-like adversarial patches to fool autonomous-driving systems. Recently, adversarial patches have been used to attack automatic checkout systems \cite{wang2021universal} and surveillance cameras \cite{thys2019fooling}. 

In this paper, we aim to generate an adversarial perturbation $\mathbf \delta$, constrained to a localized patch, to fool the crowd counting model $f_{\Theta}$ for wrong predictions. Specifically, given the crowd counting model $f_{\Theta}$, we generate adversarial patch perturbation $\mathbf \delta$ by maximizing the model loss as
\begin{equation}
\arg\max_{\delta}{||f_{\Theta}(\mathbf x_{adv})-\mathbf I||^2_2},
\end{equation}
where an adversarial example $\mathbf x_{adv}$ is composed of a clean image $\mathbf x$, an additive adversarial patch perturbation $\mathbf \delta$ $\in \mathbb{R}^z$\textcolor{red}{,} and a location mask $M$ $\in$ \{0,1\}$^n$. It can be formulated as
\begin{equation}
\mathbf x_{adv} = (1- M)\odot \mathbf x + M \odot \mathbf \delta,
\label{eqn:xadv}
\end{equation}
where $\odot$ is the element-wise multiplication. 

In the crowd counting scenario, rare attempts have been made to perform adversarial attacks. \cite{liu2019using} generated adversarial perturbations using FGSM and studied the defense against them in the digital world. \cite{wu2021towards} proposed the first and the only method APAM for adopting adversarial patches to attack crowd counting models. It aims to directly fit the target density map values, which are many times larger than the ground truth. However, it will overfit the specific model perception and thus tend to fall into local minima. For example, the multiplicative will not work for regions where the density value is 0. Through experiments, we found it fails to generate adversarial examples with strong transferability, which shows limited abilities for evaluating the black-box crowd counting models in practice. Instead, we do not constrain the target values, that is, they are expected to be infinitely far from the ground truth, and utilize scale and position perceptual properties to generate strongly transferable adversarial patches.

\section{Threat Model}




In this section, we aim to give a detailed description correlated to our proposed perceptual adversarial patches from several aspects, \ie, the possible attacking scenarios, the detailed attacking goal, the constraints to attackers, and the capabilities of attackers, therefore better benefiting the understanding of the proposed method at a practical level. 
\subsection{{Possible Attacking Scenarios}}
As for adversarial attacking tasks, one of the most important questions that should be answered is whether they are practical or not. More precisely, the existence of the potential threats or benefits associated with the attacking method decides its value and significance. 

When it comes to our proposed perceptual adversarial patches, we claim that they are applicable to multiple crowd-counting correlated scenarios, such as crowd monitoring in a particular place, population warning in a traffic scenario, and other similar scenarios. Note that we generate the adversarial patches with printing papers in this paper. However, besides utilizing paper patches as attacking vectors, we can also perform attacks by printing those adversarial textures on slogans, as shown in Figure \ref{fig:app-intro} (a) and Figure \ref{fig:phy-ours}, which strongly indicates the diverse attacking pathways of this novel perceptual adversarial patch generation framework.

\subsection{{Detailed Attacking Goal}}
Overall, we consider generating adversarial patches to attack density-map-estimation-based crowd counting models. As mentioned in Section \ref{sec:background}, given a crowd counting model $f_{\Theta}$ that takes an image $\mathbf x$ as input, attackers aim to mislead $f_{\Theta}$ into making wrong predictions, therefore outputting an inaccurate density map far from the ground-truth density map. And the achievement of this goal depends on the design of the adversarial patches.

Further, there are two directions for misleading the crowd counting model $f_{\Theta}$ into wrong density maps. One is leading it to output more crowd counting values, \ie, to increase the predicted crowd numbers. Another is leading it to output fewer crowd counting values, \ie, to decrease the predicted crowd numbers. In this paper, we respectively investigate both the increasing approach and decreasing approach to comprehensively demonstrate the strong attacking ability of the proposed perceptual adversarial patch generation framework. The experimental results of the increasing approach are mainly shown in Section \ref{digital-attack} and \ref{sec:phyattack} and those of the decreasing approach are shown in Section \ref{discussion}. 

\begin{figure*}[!ht]
\begin{center}
\includegraphics[width=1.0\linewidth]{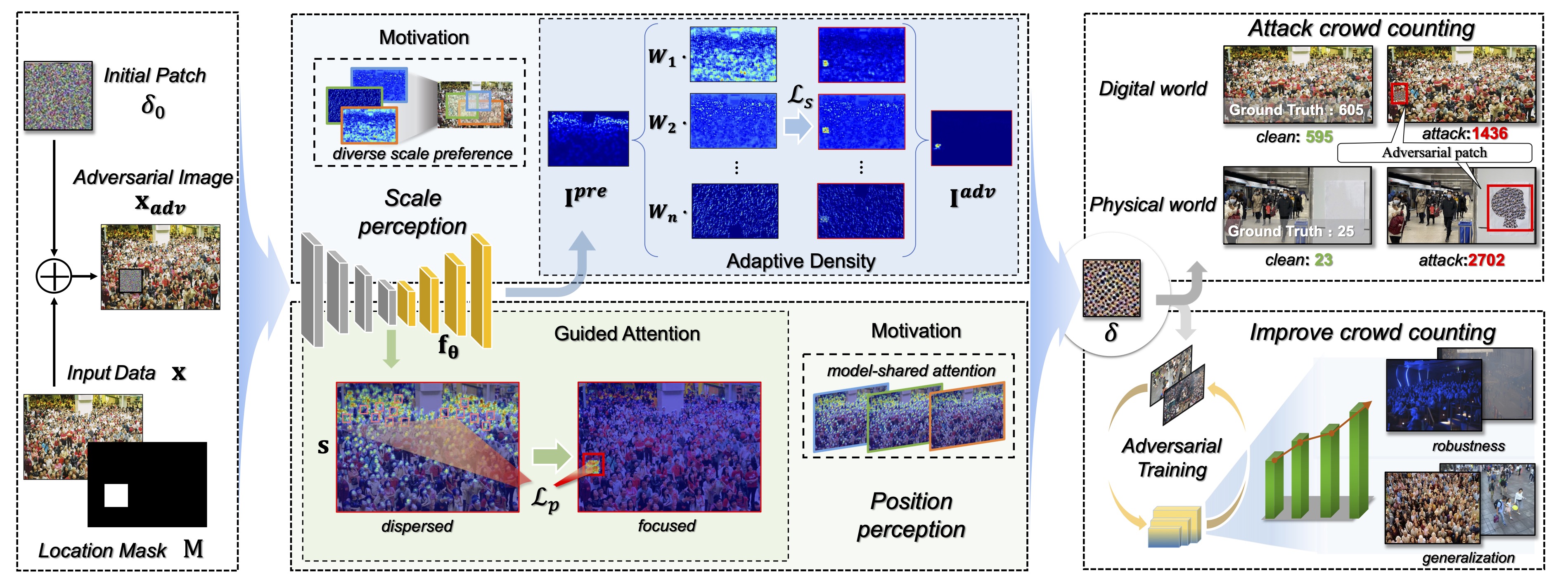}
\end{center} 
\vskip -3ex
\caption{Perceptual Adversarial Patch (PAP) generation framework. The adversarial patch is updated by jointly optimizing the scale perception loss $\mathcal L_s$ based on adaptive density and the position perception loss $\mathcal L_p$ based on guided attention. Finally, PAP can successfully attack the crowd counting models and further improve the model performance through adversarial training.} 
\label{fig:approach}
\vskip -2ex
\end{figure*}

\subsection{{Attackers' Constraint}}
With full consideration of the crowd counting scenarios in the physical world, we have to take note of the complexity and dynamics of the real-world conditions, such as unknown model architectures, unknown parameters, and unknown crowd densities. Therefore, it is necessary for us to simulate the real conditions as far as possible. To this end, we consider the more comprehensive experimental conditions, which consist of white-box and black-box settings.

In white-box experiments, we take target models, \ie, the models that might be attacked, as source models during the training process. That is, the attackers could totally access the model information, including middle-layer features, output density maps, gradient information, \etc. In black-box settings, we impose the strictest constraint on attackers, which means that the attackers could acquire little knowledge about the target models. To be more practical, we consider the transfer-based attack. More precisely, we just generate adversarial patches based on a certain model and then perform attacks without any other additional actions, \eg, fine-tuning or query. Therefore, an essential guarantee for successful attacks is strong inter-model transferability. Based on that, we could guarantee that all the information of target models is unavailable to the attackers in black-box settings, which helps us to conduct the strictest measures for simulating the physical scenarios. Besides, since we aim to generate adversarial patches, it is important for us to constrain the perturbed ratio, \ie, the ratio of perturbed pixels to all pixels. In this paper, we only perturb $81\times81$ pixels (\ie, nearly 0.83\% of a certain image) and constrain their shapes by the mask $M$ in Equation (\ref{eqn:xadv}).

\subsection{{Attackers' Capability}}
Most attackers who perform typical adversarial attacks (\ie, adversarial examples) could be classified into two categories:, adversarial perturbations and adversarial patches. The adversarial perturbations always have an invisible appearance to human beings, whereas they also show weak attacking ability in real scenarios due to the domain gap during re-sampling. Therefore, in this paper, we aim to generate adversarial patches, which are confined to patch-like textures without $\epsilon-$ball constraint. These kinds of adversarial noises could be more threatening in our crowd counting tasks, especially in the physical world. 

As for the attack workflow, we basically follow a classic attacking paradigm. Specifically, an attacker needs to first generate an adversarial patch by training it on certain datasets. Then, one can produce these adversarial textures by printing them out in the physical world. Also, adversaries could clip or shear the printed adversarial patches into particular shapes, such as the ``human-head-like patch'' that is shown in Figure \ref{fig:app-intro} (a). Finally, attackers could simply stick the handled adversarial patches into a target object and attack the deployed model in the corresponding scenarios. To sum up, the attacking paradigm of attacking with our perceptual adversarial patches can be simply described as a ``generating-producing-processing-attacking'' approach.

\section{Perceptual Adversarial Patch Generation Framework}

\subsection{Overview}
Existing studies reveal that the model-invariant characteristics largely influence the transferability of attacks \cite{dong2019evading,lennon2021patch}. Thus, we aim to find the model-shared characteristics that highly influence model performance and then learn model-invariant features from them to generate transferable adversarial patches across models. Driven by this belief, we propose the Perceptual Adversarial Patch (PAP) generation framework by introducing adaptive density and guided attention to help adversarial patches exploit the model's intrinsic perceptual characteristics, \ie, scale perception and position perception, so as to capture the model-invariant features. Thus, our generated adversarial patches could enjoy better transfer attacking abilities. The overall framework is shown in Figure \ref{fig:approach}.

\subsection{Scale Perception via Adaptive Density}

\cite{zhang2016single} illustrated that the variation of crowd scales highly challenges the design and performance of crowd counting models. It is difficult for a single model to well recognize crowd features at all scales, and different models, more or less, tend to show different crowd scale perception preferences. Therefore, capturing the scale-invariant features could benefit the adversarial patches for better adaptation to models with different crowd scale perceptions, resulting in stronger transferability. In order to achieve the objective, we introduce the adaptive density during patch generation to dynamically adjust the contribution of features with different scales.

Simply generating perturbations via Eqn (\ref{eqn:xadv}) will spontaneously capture features that overfit the specific scale perception of the source model, which leads to weak transferability. Therefore, we aim to enhance its scale capture among different models, especially scales that the source model does not perceive well. Given an input image $\mathbf x$, we generate the ground truth density map $\mathbf I$ by the geometry-adaptive kernels for the highly congested scenes, following the method in \cite{zhang2016single}. The geometry-adaptive kernel is defined as:
\begin{equation}
F(\mathbf{x}) = \sum_{i=1}^{N}{\updelta(\mathbf{x} - \mathbf{x}_i) \times G_{\sigma_i}(\mathbf{x})} \ with\  {\sigma_i}=\beta \overline{d_i},
\label{eqn:kernel}
\end{equation}
For the input image $\mathbf{x}$, if there is a head at the pixel $\mathbf {x}_i$, it can be represented as a delta function ${\updelta(\mathbf{x} - \mathbf{x}_i)}$. $N$ is the number of heads contained in $\mathbf{x}$. Then we convolve it with a Gaussian kernel with the standard deviation ${\sigma_i}$, where $\overline{d_i}$ indicates the average distance of $k$ nearest neighbors. In practice, we set $\beta=0.3$ and $k=3$, following the configuration in \cite{zhang2016single}. For the sparse scenes, we use a constant $\sigma=15$ to blur all the annotations.

By smoothing each head annotation with a Gaussian kernel, the ground truth density map $\mathbf I$ considers the spatial distribution of all input images and thus contains the full crowd scale information of the scenario. We, therefore, propose the density weights matrix $W$ as follows:
\begin{equation}
W = Sig(\mathbf I - {f_{\Theta}}(\mathbf{x}_{adv})),
\label{eqn:W}
\end{equation}
where $Sig(\cdot)=1/(1+e^{-(\cdot)})$ denotes the $\emph{Sigmoid}$ function. Apparently, for the crowd region with specific scales that are included in the ground truth while not perceived by the source model based on the density map, $W$ will be increased to a higher value. In other words, the crowd scales perceived weakly by the model will be granted higher weights, which will help the patches to better adapt to them.

Since our goal is to mislead the model to wrong predictions, we intuitively ought to force the model to recognize the adversarial patches as crowds to a large extent. Therefore, based on the weighted predicted density map, we introduce the scale perception loss $\mathcal L_s$ as follows:
\begin{equation}
\mathcal L_s = \sum_{i,j}{{W}_{i,j}{f_{\Theta}}_{i,j}(\mathbf{x}_{adv})},
\label{eqn:lsloss}
\end{equation}
where ${W}_{i,j}$ and ${f_{\Theta}}_{i,j}$ are the value at $(i,j)$ of the ${W}$ and the predicted density map. Through the density weights, our adversarial patch could capture the scale-invariant features and better adapt to the scale perceptions of different models, which will benefit the transferability.

\begin{algorithm}[!ht]
\caption{Perceptual Adversarial Patch Generation}
\label{alg:alg1}
\begin{algorithmic}
\renewcommand{\algorithmicrequire}{\textbf{Input:}}
\renewcommand{\algorithmicensure}{\textbf{Output:}}
\REQUIRE Initial patch noise $\delta_0$, image set $\mathbb X=\{\mathbf x_i|i=1,\cdots,n\}$, target model $f_{\Theta}$, hyperparameters $\lambda,\alpha, T$
\ENSURE Adversarial patch perturbation $\delta$
\STATE generate the ground truth density map set $\mathbb I=\{\mathbf I_i|i=1,\cdots,n\}$
\STATE Initialize $\delta \leftarrow \delta_0$
\FOR{the number of epochs}
\STATE select $minibatch$ images from $\mathbb X$
\FOR{$m = n/minibatch$ steps}
\STATE randomly generate a location mask $M$
\STATE $t \leftarrow 0$, $\delta^t \leftarrow \delta$
\FOR{$t < T$}
\STATE \#generate adversarial examples: \\ $x_{adv} \leftarrow (1- M)\odot \mathbf x + M \odot \mathbf \delta^t$
\STATE \#clip to the normal range: \\ $x_{adv} \leftarrow clip(x_{adv},[0,1])$
\STATE \#get the density weights matrix: \\ $W \leftarrow Sig(\mathbf I - {f_{\Theta}}(\mathbf{x}_{adv}))$
\STATE \#get the attention map: \\ $S \leftarrow \mathcal{A}(\mathbf{x}_{adv},f_{\Theta})$
\STATE \#compute the loss: \\ $\mathcal{L}_{s} \leftarrow \sum_{i,j}{{W}_{i,j}{f_{\Theta}}_{i,j}(\mathbf{x}_{adv})}$, $\mathcal{L}_{p} \leftarrow \sum_{i,j} S_{i,j}$, \\
$\mathcal{L}_{total} \leftarrow \mathcal{L}_{s} + \lambda\mathcal{L}_{p}$
\STATE \#update the adversarial perturbation: \\ $\delta^{t+1} \leftarrow \delta^t - \alpha \cdot \frac{\partial \mathcal{L}_{total}}{\partial \delta^t}$
\ENDFOR
\STATE $\delta \leftarrow \delta^{t}$
\ENDFOR
\ENDFOR
\end{algorithmic}
\end{algorithm}

\subsection{Position Perception via Guided Attention}
Previous work reveals that different models share similar positional perceptions towards the same image \cite{wang2021dual}. As for crowd counting models, we find that they also have similar spatially dispersed attention patterns at the same crowd positions. Therefore, we disturb the position perception of the target model by attracting the model-shared attention patterns to the adversarial patch region through salient map aggregation. In this way, the generated adversarial patches can capture the position-invariant features and perform better transferable attacks.

In particular, several visual attention mechanisms \cite{zhou2016learning,selvaraju2017grad,chattopadhay2018grad} have been proposed to explain deep learning behaviors. Grad-CAM \cite{selvaraju2017grad} is a class-discriminative localization technique that can generate visual explanations from any CNN-based network. Given an input image and a model, the method could produce a salient map with hot regions where the pixel values are higher. It reveals that the model will pay more attention to the regions which are meaningful to the final predictions. When it comes to crowd counting tasks, the density map of a certain image to be predicted also shows significant differences among diverse sub-parts, which inspires us to regard this observation as the density perception of models. Therefore, by introducing the idea of the Grad-CAM, we elaborately design to calculate the density-guided attention map for helping the generated adversarial patches to disturb the position perception and capture the position-invariant features in turn, leading to better transferable attacks. Specifically, given the image $\mathbf x_{adv}$ and a target model $f_{\Theta}$, we compute the attention map $\mathbf{S}$ by introducing a density attention module $\mathcal{A}$ as:
\begin{equation}
    \begin{aligned}
        {S}&=\mathcal{A}(\mathbf{x}_{adv},f_{\Theta}) \\
        &= \emph{ReLU}(\frac{1}{Z}\sum_{i,j,k}\frac{\partial C}{\partial A^{k}_{ij}}\cdot A^{k}),
    \end{aligned}
\label{eqn:attention}
\end{equation}
where $C=\sum_{i,j}{{f_{\Theta}}_{i,j}(\mathbf{x}_{adv})}$, $A^{k}_{ij}$ is the pixel value at position $(i,j)$ of the $k$th feature map, $\emph{ReLU}(\cdot)=max(0,\cdot)$ denotes the $\emph{ReLU}$ function, and $Z$ is for global average pooling. The conventional Grad-CAM computes the gradients of the scores for specific classes, while we utilize the summary of density values, \ie, the people number, to obtain the gradients. Thus, we can generate the salient map, which can be used to explain the decision basis of the crowd counting models.

In order to successfully attack a crowd counting model, we draw the model's attention to our adversarial patches and thus distract it from other crowds. We introduce the position perception loss as follows, which directly increases the attention values in the patch region:
\begin{equation}
\begin{split}
\mathcal L_p = \sum_{i,j} S_{i,j}(\mathbf{x}_{adv}),
\label{eqn:laloss}
\end{split}
\end{equation}
where $S_{i,j}$ is the pixel value at $(i,j)$ of the attention map. Thus, different models with similar salient attention areas will focus on the adversarial patches and make the wrong predictions.

\subsection{Overall Optimization}
In this section, we aim to give a brief operation procedure of our proposed perceptual adversarial patches, therefore establishing an integral cognition of the novel crowd counting attacking method for readers.

In general, given a target model $f_{\Theta}$, hyperparameters $\lambda$, $\alpha$, $T$, dataset $\mathbb{X}$, and initial patch noise $\delta_0$, we generate the adversarial patches by jointly optimizing the scale perception loss $\mathcal{L}_{s}$ and position perception loss $\mathcal{L}_{p}$. The overall optimization for generating the transferable adversarial patches $\delta$ could be formulated by the following equation: 
\begin{equation}
\label{eqn:total-loss}
    \begin{split}
     \mathop{\arg\max}_{\delta}\mathcal{L}_{s} + \lambda\mathcal{L}_{p},
    \end{split}
\end{equation}
where $\lambda$ controls the contributions of each term. Specifically, we totally employ the gradient-based iteration algorithm to optimize our adversarial patches. In each iteration, we first generate adversarial examples with an initial adversarial patch at a random position; then we conduct the forward pass to obtain the predicted density map; next, we derive the density weights matrix $W$ and attention map $S$, and subsequently compute the scale perception loss and position perception loss; finally, we update the adversarial patch through the back-propagation algorithm \cite{rumelhart1986learning} to lead the model to wrong density predictions and enhance the model-shared attention towards the patch region. By strictly conducting the described operation procedures, we can efficiently generate adversarial patches with strong transferability by exploiting the model-shared perceptual features, \ie, scale perception and position perception. The overall detailed training algorithm can be described as Algorithm \ref{alg:alg1}.

\begin{table*}[!ht]
\scriptsize
\caption{Results of different patch attacks for crowd counting on the Shanghai Tech dataset. The results on the diagonal are in white-box settings while the others are in black-box settings. Higher MAE and MSE values indicate a stronger attack.}
\centering
\setlength{\tabcolsep}{5mm}
\begin{tabular}{cccccccc}
\toprule
\multicolumn{2}{c}{\textbf{MAE / MSE}}&\multicolumn{6}{c}{\textbf{Target Model}}\\ 
\midrule
\textbf{{Source model}} & \textbf{Method} & MCNN                   & CSRNet & CAN & BL & DM-Count & SASNet\\
\midrule
\multicolumn{2}{c}{Part A}\\
\midrule
\rowcolor{gray!20}
\multicolumn{2}{c}{Clean}
&{108.0} / {165.0}         
&{ 67.0} / {105.2}           
&{59.9} / {94.1}           
&{61.8} / {94.1}             
&{58.2} / {93.2}            
&{52.8} / {86.2}           
\\
\multirow{2}{*}{MCNN} 
& APAM   
& \cellcolor{blue!20}{317.7} / {378.5}            
& { 68.9} / {107.2}           
& {62.9} / {99.3}
& {63.7} / {96.2}             
& {61.4} / {96.4}           
& {54.2} / {87.9}           
\\ \multirow{2}{*}{} 
& \textbf{Ours}   
& \cellcolor{blue!20}\textbf{908.7} / \textbf{989.1} 
& \textbf{ 69.6} / \textbf{109.3}  
& \textbf{ 62.9} / \textbf{101.3}   
& \textbf{64.1} / \textbf{96.9}     
& \textbf{62.0} / \textbf{96.8}    
& \textbf{54.3} / \textbf{89.9}   
\\
\multirow{2}{*}{CSRNet} 
& APAM   
& {124.6} / {174.8}            
& \cellcolor{blue!20}{ 312.5} / {396.4}           
& { 71.3} / {100.9}           
& {250.8} / {265.2}             
& {131.1} / {151.4}            
& {53.7} / {87.9}          
\\ \multirow{2}{*}{} 
& \textbf{Ours}   
& \textbf{147.0} / \textbf{190.8}   
& \cellcolor{blue!20}\textbf{568.4} / \textbf{613.8} 
& \textbf{212.5} / \textbf{242.8} 
& \textbf{388.1} / \textbf{401.4}   
& \textbf{249.1} / \textbf{263.6}  
& \textbf{56.3} / \textbf{89.5}   
\\
\multirow{2}{*}{CAN} 
& APAM   
& {133.5} / {180.5}            
& { 99.9} / {128.8}           
& \cellcolor{blue!20}{386.5} / {500.0}           
& {272.1} / {285.6}             
& {144.6} / {165.1}            
& {53.9} / {88.5}           
\\ \multirow{2}{*}{} 
& \textbf{Ours}   
& \textbf{147.6} / \textbf{191.5}   
& \textbf{321.0} / \textbf{341.7} 
& \cellcolor{blue!20}\textbf{513.3} / \textbf{545.3} 
& \textbf{412.7} / \textbf{424.8}   
& \textbf{218.8} / \textbf{233.9}  
& \textbf{56.2} / \textbf{88.8}   
\\
\multirow{2}{*}{BL}              
& APAM   
& {115.1} / {168.6}             
& { 69.9} / {106.2}           
& {62.9} / {97.0}           
& \cellcolor{blue!20}{138.3} / {160.0}             
& {103.5} / {130.3}            
& {54.0} / {88.0}  
\\ \multirow{2}{*}{} 
& \textbf{Ours}   
& \textbf{119.0} / \textbf{170.7}  
& \textbf{ 79.6} / \textbf{111.5}  
& \textbf{ 73.1} / \textbf{106.6}   
& \cellcolor{blue!20}\textbf{1090.9} / \textbf{1171.6} 
& \textbf{519.7} / \textbf{541.6}  
& \textbf{54.5} / \textbf{90.0}   
\\
\multirow{2}{*}{DM-Count}  
& APAM   
& {115.0} / {169.0}            
& { 69.6} / {107.3}           
& {62.2} / {98.4}           
& { 81.2} / {113.7}             
& \cellcolor{blue!20}{112.5} / {147.5}            
& {54.1} / {88.1}           
\\ \multirow{2}{*}{} 
& \textbf{Ours}   
& \textbf{115.6} / \textbf{169.6}   
& \textbf{ 87.6} / \textbf{119.3}  
& \textbf{ 82.8} / \textbf{115.0}   
& \textbf{747.5} / \textbf{793.6}   
& \cellcolor{blue!20}\textbf{751.3} / \textbf{784.9} 
& \textbf{56.5} / \textbf{90.8}   
\\
\multirow{2}{*}{SASNet} 
& APAM   
& {110.5} / {169.1}            
& { 68.3} / {108.4}           
& {61.7} / {99.3}  
& {63.6} / {98.4}             
& {62.3} / {99.7}           
& \cellcolor{blue!20}{58.8} / {99.8}           
\\ \multirow{2}{*}{} 
& \textbf{Ours}   
& \textbf{112.8} / \textbf{169.4}   
& \textbf{ 69.5} / \textbf{109.0}  
& \textbf{ 62.6} / \textbf{100.3}      
& \textbf{69.9} / \textbf{99.8}    
& \textbf{ 75.1} / \textbf{105.4}   
& \cellcolor{blue!20}\textbf{200.0} / \textbf{220.3} 
\\\midrule
\multicolumn{2}{c}{Part B}\\
\midrule
\rowcolor{gray!20}
\multicolumn{2}{c}{Clean} 
& {28.3} / {38.7}         
& { 9.2} / {14.7}           
& { 7.5} / {11.9}           
& { 7.6} / {12.0}             
& { 7.3} / {11.8}            
& {6.4} / {9.9}           
\\ 
\multirow{2}{*}{MCNN} 
& APAM   
& \cellcolor{blue!20}{29.1} / {39.7}            
& {10.8} / {16.0}           
& { 8.4} / {12.5}
& { 7.7} / {12.8}             
& { 7.7} / {12.1}           
& { 7.0} / {10.6}           
\\ \multirow{2}{*}{} 
& \textbf{Ours}   
& \cellcolor{blue!20}\textbf{442.6} / \textbf{445.2} 
& \textbf{11.0} / \textbf{16.9}  
& \textbf{26.6} / \textbf{28.8}   
& \textbf{ 7.8} / \textbf{13.3}     
& \textbf{ 7.7} / \textbf{12.5}    
& \textbf{ 7.2} / \textbf{10.9}   
\\ 
\multirow{2}{*}{CSRNet} 
& APAM   
& {66.7} / {72.8}            
& \cellcolor{blue!20}{83.3} / {87.4}           
& {21.7} / {24.1}           
& {21.4} / {25.0}             
& { 8.3} / {13.0}            
& { 6.9} / {10.5}          
\\ \multirow{2}{*}{} 
& \textbf{Ours}   
& \textbf{162.8} / \textbf{167.9}   
& \cellcolor{blue!20}\textbf{948.2} / \textbf{961.3} 
& \textbf{112.0} / \textbf{113.8} 
& \textbf{48.4} / \textbf{50.7}   
& \textbf{27.4} / \textbf{37.3}  
& \textbf{ 7.2} / \textbf{10.9}   
\\ 
\multirow{2}{*}{CAN} 
& APAM   
& {28.9} / {42.3}            
& {10.6} / {15.9}           
& \cellcolor{blue!20}{ 8.3} / {12.5}           
& { 7.7} / {12.9}             
& { 7.6} / {12.0}            
& { 6.9} / {10.5}           
\\ \multirow{2}{*}{} 
& \textbf{Ours}   
& \textbf{29.0} / \textbf{42.7}   
& \textbf{11.2} / \textbf{17.3} 
& \cellcolor{blue!20}\textbf{37.0} / \textbf{39.4} 
& \textbf{ 7.8} / \textbf{13.3}   
& \textbf{ 7.8} / \textbf{12.6}  
& \textbf{ 7.2} / \textbf{11.0}   
\\ 
\multirow{2}{*}{BL}              
& APAM   
& {53.5} / {59.6}             
& {10.9} / {16.1}           
& { 9.8} / {13.5}           
& \cellcolor{blue!20}{13.9} / {18.8}             
& { 7.7} / {12.1}            
& { 6.9} / {10.5}  
\\ \multirow{2}{*}{} 
& \textbf{Ours}   
& \textbf{69.3} / \textbf{75.0}  
& \textbf{28.4} / \textbf{32.9}  
& \textbf{81.0} / \textbf{82.4}   
& \cellcolor{blue!20}\textbf{146.5} / \textbf{147.5} 
& \textbf{58.3} / \textbf{69.9}  
& \textbf{ 7.2} / \textbf{10.9}   
\\ 
\multirow{2}{*}{DM-Count}  
& APAM   
& {47.9} / {54.4}            
& {15.6} / {20.5}           
& {12.0} / {15.3}           
& { 7.9} / {13.1}             
& \cellcolor{blue!20}{ 9.8} / {16.7}            
& { 6.9} / {10.5}           
\\ \multirow{2}{*}{} 
& \textbf{Ours}   
& \textbf{65.2} / \textbf{71.1}   
& \textbf{162.6} / \textbf{167.0}  
& \textbf{72.7} / \textbf{74.2}   
& \textbf{83.6} / \textbf{84.9}   
& \cellcolor{blue!20}\textbf{295.6} / \textbf{297.1} 
& \textbf{ 7.2} / \textbf{10.9}   
\\ 
\multirow{2}{*}{SASNet} 
& APAM   
& {33.9} / {41.8}            
& {10.4} / {16.5}           
& {10.7} / {14.7}  
& { 9.4} / {14.6}             
& { 7.7} / {12.7}           
& \cellcolor{blue!20}{ 7.0} / {10.9}           
\\ \multirow{2}{*}{} 
& \textbf{Ours}   
& \textbf{ 95.3} / \textbf{101.2}   
& \textbf{177.2} / \textbf{180.5}  
& \textbf{114.9} / \textbf{116.2}      
& \textbf{ 39.7} / \textbf{42.3}    
& \textbf{ 10.4} / \textbf{16.1}   
& \cellcolor{blue!20}\textbf{293.0} / \textbf{295.3} 
\\ \bottomrule
\end{tabular}
\label{tab:black-box attacks}
\end{table*}

\section{Evaluation of PAP Attack}
In this section, we first outline the experimental settings and then illustrate the effectiveness of our proposed attacking method by thorough evaluations in both the digital and physical world. Finally, we provide some additional discussions.

\subsection{Experimental Settings}\label{attack-setting}

\paragraph{\textbf{Datasets.}} Following \cite{wu2021towards}, we conduct experiments on the Shanghai Tech dataset \cite{zhang2016single}, a commonly used large-scale crowd counting dataset. It consists of 1198 annotated crowd images with 330,165 annotated people. The dataset is divided into Part A and Part B. Part A contains 300 samples for training and 182 samples for testing, where images were collected from the Internet. Part B contains 400 samples for training and 316 samples for testing, where images were collected on the busy streets of Shanghai.

\paragraph{\textbf{Target models.}} We employ six commonly-used and SOTA density-map-estimation-based  crowd counting models to attack: MCNN \cite{zhang2016single}, CSRNet \cite{li2018csrnet}, CAN \cite{liu2019context}, BL \cite{ma2019bayesian}, DM-Count \cite{wang2020distribution}, and SASNet \cite{song2021choose}. Among them, \cite{zhang2016single,song2021choose} are multi-column methods while the others are single-column methods. In addition to the vanilla model, we also conduct attacks towards the empirical defensive method based on adversarial training \cite{madry2017towards} and certified defense against crowd counting based on randomized ablation \cite{wu2021towards} \footnote{Implementation details and results can be found in appendix section B}.

\paragraph{\textbf{Evaluation metrics.}}
We use the widely-used crowd counting metrics Mean Absolute Error (MAE) and Mean Squared Error (MSE) following \cite{li2018csrnet} for evaluation, which are defined as:
\begin{equation}
\begin{split}
MAE = \frac{1}{N}\sum_{i=1}^N |C_i-C_i^{GT}|,\ MSE = \sqrt{\frac{1}{N}\sum_{i=1}^N |C_i-C_i^{GT}|^2},
\label{eqn:mae}
\end{split}
\end{equation}
where N is the size of the test set, $C_i^{GT}$ is the ground truth of counting and $C_i$ represents the estimated count. For attacks, higher MAE and MSE values indicate stronger adversarial attacks.

\paragraph{\textbf{Baselines.}} We compare with the only adversarial patch generation method for crowd counting, \ie, APAM \cite{wu2021towards}. We use the officially released codes and keep the same settings (size, shape, position, \etc) for fair comparisons. Besides, we also compare with several plug-and-play transferable attacks (MIGM \cite{dong2018boosting}, NIGM \cite{lin2019nesterov}, TI-NIGM \cite{dong2019evading}, NAA \cite{zhang2022improving}) and ensemble-based attacks (Avg-Dens, MGAA \cite{yuan2021meta}).

\begin{figure}[!ht]
\begin{center}
\includegraphics[width=0.9\linewidth]{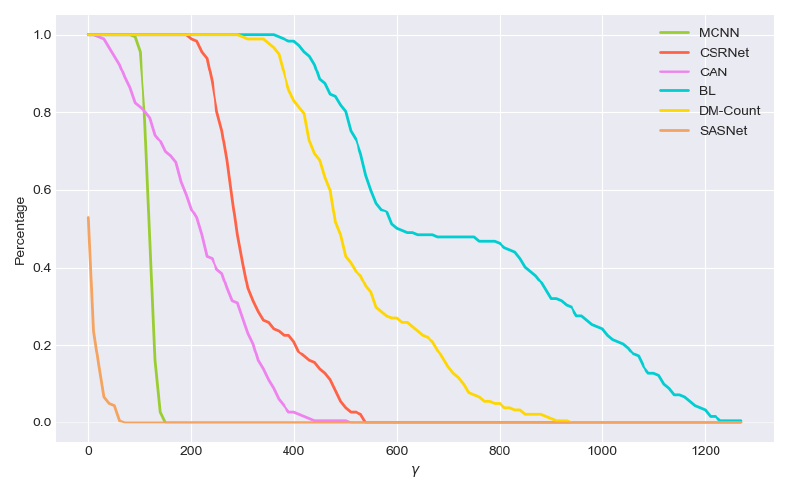}
\end{center} 
\vskip -3ex
\caption{Percentage of overestimation of crowd counting models under black-box PAP attack aiming to increase the crowd numbers on Shanghai Tech Part A.}
\label{fig:4}
\vskip -3ex
\end{figure}  

\paragraph{\textbf{Implementation details.}}
We randomly initialize a square adversarial patch with a fixed size and conduct training with batch size 1 by $T = 25$ iterations every epoch with an attack step size $\alpha$ of 0.01, and a maximum of 2 epochs. The position and orientation of the patch are randomly chosen, which makes our adversarial patches able to universally attack all scenes. We set the position perception loss weight $\lambda$ as 0.01. We give a detailed discussion related to the effect of $\lambda$ and different patch shapes in Section \ref{discussion}. All of our codes are implemented in PyTorch. We conduct all experiments on an NVIDIA Tesla V100 GPU cluster.

\begin{table*}[!ht]
\scriptsize
\caption{Black-box results of various methods designed for transferable attacks on the Shanghai Tech dataset Part A. Higher MAE and MSE values indicate a stronger attack.}
\centering
\setlength{\tabcolsep}{7mm}
\begin{tabular}{ccccccc}
\toprule
{\textbf{MAE / MSE}}&\multicolumn{6}{c}{\textbf{Target Model}}\\ 
\midrule
\textbf{Method} & MCNN  & CSRNet & CAN & BL & DM-Count & SASNet\\
\midrule
MIGM   
& {145.2} / {188.9}            
& {255.6} / {273.9}           
& {104.2} / {128.8}
& {703.9} / {713.2}             
& {491.9} / {504.9}           
& {55.9 } / {90.3 }           
\\
NIGM   
& {146.9} / {190.2} 
& {267.2} / {285.3}  
& { 87.4} / {110.7}   
& {701.0} / {731.2}     
& {511.7} / {526.5}    
& {55.8 } / {90.0 }   
\\ 
TI-NIGM  
& {111.2} / {169.3}            
& { 74.5} / {109.8}           
& { 65.4} / {98.1 }           
& { 75.8} / {104.3}             
& { 77.9} / {107.3}            
& { 54.5} / {89.9 }          
\\
NAA   
& {139.3} / {184.5}   
& {257.3} / {276.9} 
& { 93.8} / {120.4} 
& {737.7} / {746.0}   
& {508.9} / {522.1}  
& {56.4 } / {90.8 }   
\\
Avg-Dens 
& {136.0} / {182.6}            
& {220.2} / {238.9}           
& {128.8} / {158.5}           
& {650.7} / {660.4}             
& {403.3} / {417.4}            
& \textbf{58.1} / \textbf{90.9}           
\\
MGAA  
& {135.3} / {181.2}   
& {205.4} / {225.6} 
& {118.1} / {148.7} 
& {685.5} / {695.7}   
& {416.2} / {430.8}  
& {56.5} / {90.3}   
\\
\textbf{Ours}   
& \textbf{147.6} / \textbf{191.5}  
& \textbf{321.0} / \textbf{341.7}  
& \textbf{212.5} / \textbf{242.8}   
& \textbf{747.5} / \textbf{793.6} 
& \textbf{519.7} / \textbf{541.6}  
& {56.5} / {90.8}
\\ \bottomrule
\end{tabular}
\label{tab:2}
\end{table*}

\begin{figure*}[!ht]
\begin{center}
\includegraphics[width=1.0\linewidth]{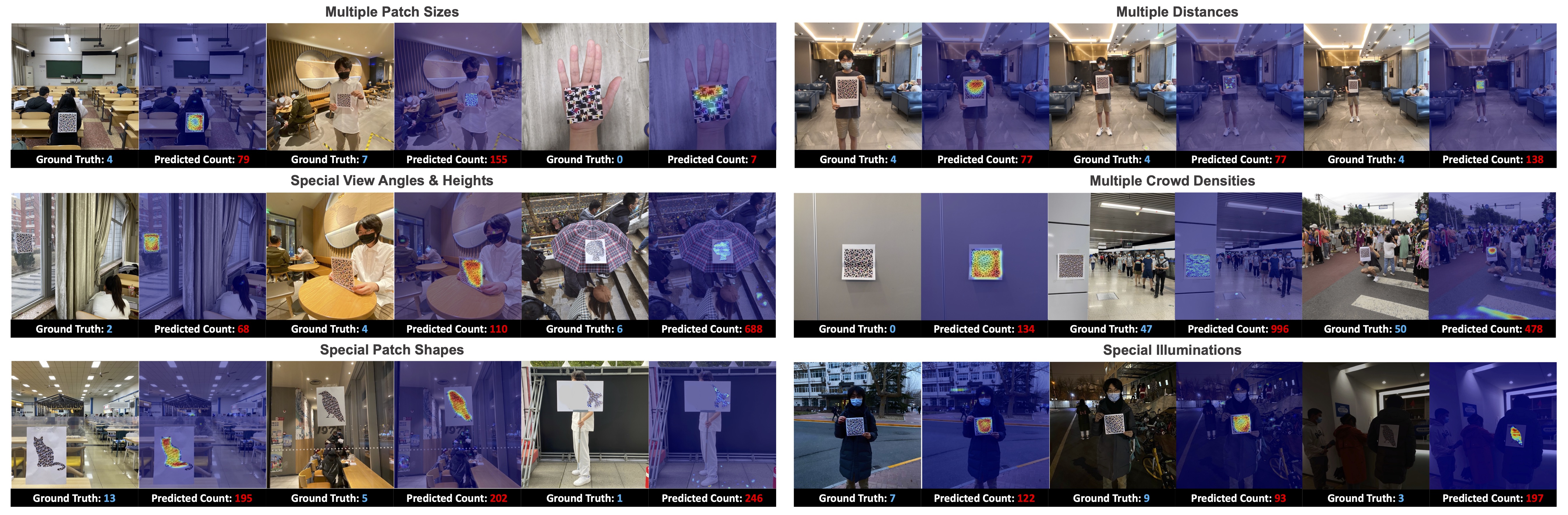}
\end{center} 
\vskip -2ex
\caption{Physical world attack in the real-world scenario. Our adversarial patches can mislead the crowd counting model under different scenes in practice.}
\label{fig:phy-ours}
\vskip -2ex
\end{figure*}

\subsection{Digital World Attack}\label{digital-attack}
We evaluate the performance of our adversarial patches in the digital world under both white-box and black-box settings. Due to the limited space, we hereby only report the results of adversarial patches with the size of $81\times81$ which only accounts for 0.83\% of the size of images in the Shanghai Tech dataset \footnote{For other patch sizes, please refer to appendix section A}. For a fair comparison with APAM, we here report the results of attacks that increase the counting number. Besides, we can also generate adversarial patches that decrease the crowd counting number and defer the results in Section \ref{discussion}. 

\paragraph{\textbf{White-box attacks.}}
For the white-box attack, we generate adversarial patches using the specific target model and perform attacks on it accordingly. As shown in Table \ref{tab:black-box attacks} (diagonal), in contrast to APAM, our method achieves higher MAE and MSE in the white-box settings on different models (up to \textbf{+1029.1 MAE} and \textbf{+1077.5 MSE} on BL). Therefore, our method is able to generate adversarial patches with a much stronger white-box attacking ability.

\paragraph{\textbf{Black-box attacks.}}
In the black-box setting, we first generate adversarial patches based on one specific source model, and then transfer the attacks to other models and test their attacking ability. Tables \ref{tab:black-box attacks} and \ref{tab:2} list the black-box attacking results with a series of patch attacks and transferable attacks. Regarding MIGM, NIGM, TI-NIGM, and NAA and Ours in Table \ref{tab:2}, we select the highest MAE/MSE of a target model from the results of five source models\footnote{Implementation details and more results can be found in appendix section C}. In Figure \ref{fig:4}, we plot the model overestimation curves with a sweep of $\gamma$. Specifically, for each target model, we consider the source model which possesses the best black-box attacking performance referring to Table \ref{tab:black-box attacks}, and calculate the percentage of the samples on which the model overestimation value, \ie, $C^{adv}_{pre}-C^{clean}_{pre}$, exceeds the $\gamma$. Through the results, we can draw some \textbf{observations:}

(1) Compared with APAM, we achieve stronger black-box attacking ability for different models and outperform it by large margins (up to \textbf{+685.7 MAE}, \textbf{+699.5 MSE} from DM-Count to BL). According to Figure \ref{fig:4}, almost all model overestimation values achieve above \textbf{100} for \textbf{80\%} of the samples.

(2) Compared with other transferable attacks, we significantly beat them except for being slightly worse than Avg-Dens on SASNet. We note that TI-NIGM has a much weaker attacking ability, which illustrates that the translation-invariant method may not be well suitable for the crowd counting task.

(3) We found that adversarial attacks could hardly transfer between multi-column (\eg, MCNN and SASNet) and single-column models. We conjecture the reasons might be those multi-column models have more complex architectures with several branches and more information redundancy \cite{li2018csrnet}. These architectures might cause the weak black-box transferability of adversarial attacks, and we leave the detailed analyses as future work.

\subsection{Physical World Attack}\label{physical-attack}
\label{sec:phyattack}

Here, we further evaluate the practical performance of our adversarial patches in the physical world, which is more challenging and meaningful. We first generated an adversarial patch using the CSRNet model and printed it. Then, we took 110 pictures with an iPhone 11 mobile phone by holding them or sticking them as a flag or poster. To prove its effectiveness in the complex real-world scenario, we took photos in various settings, including:

$\bullet$ patch sizes: resizing the generated $81\times81$ patches to [5cm$\times$5cm, 40cm$\times$40cm];

$\bullet$ distances: placing the camera [1m, 5m] away from the patch;

$\bullet$ view angles and heights: considering special view angle offsets, \eg, left or right deflection, and special view heights, \eg, top or bottom view;

$\bullet$ patch shapes: cutting the patch into the shape of a cat, bird, plane and so on;

$\bullet$ illuminations: considering special illumination conditions such as dusk and darkness;

$\bullet$ crowd densities: considering various density conditions, including scenes with no people and very congested scenarios.

For each setting, we took photos in different places (schools, cafes, subway stations, \etc.). All pictures were taken with the same patch texture, that is, there is no need to re-generate the adversarial patch for different scenes. We evaluate the performance using a black-box SoTA crowd counting model DM-Count and the error caused by our adversarial patch is able to achieve \textbf{135.4} for MAE and \textbf{178.9} for MSE. As shown in Figure \ref{fig:phy-ours}, the generated adversarial patches are quite natural in the real world and will pose safety problems when deployed in practice.

\begin{figure*}[!ht]
\begin{center}
\vskip -2ex
\includegraphics[width=1.0\linewidth]{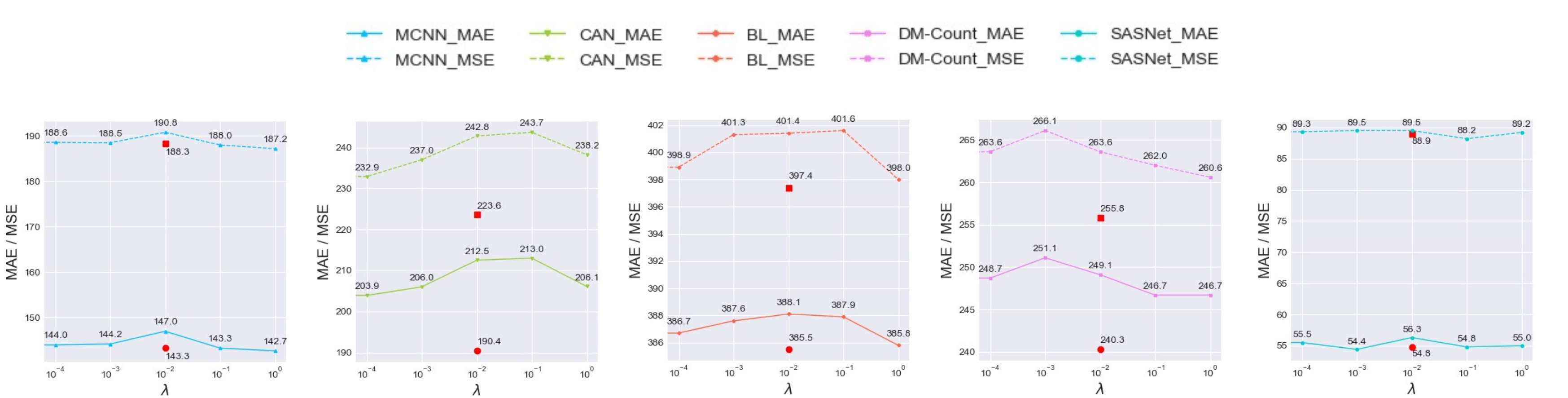}
\end{center} 
\caption{The ablation study on the influence of $\lambda$. The ``red circle" and ``red square" means the MAE and MSE when $\lambda=0$. We show the MAE/MSE for the black-box attack on Shanghai Tech Part A (based on the source model CSRNet). Higher MAE and MSE values indicate a stronger attack.}
\label{fig:ablation}
\vskip -2ex
\end{figure*}

\subsection{Discussion}\label{discussion}
In this section, we first analyze the effectiveness of the two losses. Then, we discuss the influence of different patch shapes and attacking effects across datasets. Finally, we propose the method that utilizes our PAP to maliciously decrease the predicted crowd counting numbers.

\paragraph{\textbf{The effect of the two loss functions}}
We conduct ablation studies to further investigate the contributions of scale perception and position perception, \ie, $\mathcal{L}_{s}$ and $\mathcal{L}_{p}$. Thus, we generate adversarial patches with or without these two losses from one specific model and then perform transfer attacks to other models on the Shanghai Tech Part A. Due to the limited space, we here only report the results of the source model CSRNet \footnote{More results can be found in appendix section D}.

\begin{table}[!ht]
\caption{The ablation study on loss functions. We show the MAE/MSE for the black-box attack on Shanghai Tech Part A (based on the source model CSRNet). Higher MAE and MSE values indicate a stronger attack.}
\begin{center}
\setlength{\tabcolsep}{1mm}
\scriptsize
\begin{tabular}{cccccc}
\toprule
{\textbf{Loss}}&{\textbf{MCNN}}&{\textbf{CAN}}&{\textbf{BL}}&{\textbf{DM-Count}}&{\textbf{SASNet}}\\
\midrule
{None}
&{108.0} / {165.0}
&{59.9} / {94.1}
&{61.8} / {94.1}
&{58.2} / {93.2}
&{52.8} / {86.2}\\
{$\mathcal L_s$ w/o $W$}
&{138.4} / {184.1}
&{185.4} / {216.6}
&{374.0} / {386.9}
&{239.4} / {254.9}
&{54.6} / {88.8}\\
{$\mathcal L_s$}
&{143.3} / {188.3}
&{190.4} / {223.6}
&{385.5} / {397.4}
&{240.3} / {255.8}
&{54.8} / {88.9}\\
{$\mathcal L_p$}
&{116.2} / {167.9}
&{94.6} / {121.5}
&{251.9} / {266.3}
&{152.9} / {169.8}
&{54.3} / {87.0}\\
{\textbf{$\mathcal L_s + \lambda\mathcal L_p$}}
&\textbf{147.0} / \textbf{190.8}
&\textbf{212.5} / \textbf{242.8}
&\textbf{388.1} / \textbf{401.4}
&\textbf{249.1} / \textbf{263.6}
&\textbf{56.3} / \textbf{89.5}\\
\bottomrule
\end{tabular}
\end{center}
\vskip -2ex
\label{tab:ablation}
\end{table}

As shown in Table \ref{tab:ablation}, compared with the clean scenario, the MAE and MSE values for attacking all target models increase after adding the perception loss $\mathcal{L}_{s}$. This proves that the patch could successfully mislead the model under the drive of the loss item. Furthermore, we removed the adaptive density weight matrix $W$ from the loss, \ie, the scale perception loss was represented only by the summary of predicted density values. We found that the MAE and MSE have a significant drop, which validates that the density weight matrix $W$ plays a key role in benefiting the transferability. Meanwhile, the transfer attacking ability is also improved after introducing the position loss $\mathcal{L}_{p}$. We achieve the highest MAE and MSE values when two modules are added ($\lambda=0.01$), which illustrates that the model scale perception and position perception are not completely independent, and the utilization of both can play a superimposed role in the attacking transferability. Thus, the above experimental results demonstrate the effectiveness of our dual perception loss for improving the transferability of attacks.

Ulteriorly, we analyze the influence of the hyperparameter $\lambda$. Refer to the above, we conduct transfer black-box attacks on the Shanghai Tech Part A and set the $\lambda$ as $0$, $10^{-4}$, $10^{-3}$, $10^{-2}$, $10^{-1}$, and $10^{0}$. As illustrated in Figure \ref{fig:ablation}, the MAE and MSE will mostly increase after introducing the position perception loss. Nevertheless, according to the results, we found that excessively small or large $\lambda$ values might cause the decline of the transfer attacking ability. Based on that, we set $\lambda$ as $10^{-2}$ during patch optimization.

\paragraph{\textbf{The influence of the patch shapes}}
We have considered several different patch shapes (\eg, cat, bird) in the physical world, and they show comparable performances. Here, we conduct additional ablations to further evaluate the influence of various patch shapes. We conduct a toy experiment on the Shanghai Tech Part A. Specifically, we clip the adversarial patches into three different shapes including circular, square, and trapezoid under similar patch sizes (0.83\% of the image size). Then we evaluate their transfer attack performance based on the source model CSRNet. As illustrated in Table \ref{tab:shape}, patches with different shapes show similar attacking results. Thus, we conclude that the shape does not affect transferability.

\begin{table}[!ht]
\caption{The ablation study on different patch shapes. We show the MAE/MSE for the black-box attack on Shanghai Tech Part A (based on the source model CSRNet). Higher MAE and MSE values indicate a stronger attack.}
\begin{center}
\setlength{\tabcolsep}{1mm}
\scriptsize
\begin{tabular}{cccccc}
\toprule
{\textbf{Shape}}&{\textbf{MCNN}}&{\textbf{CAN}}&{\textbf{BL}}&{\textbf{DM-Count}}&{\textbf{SASNet}}\\
\midrule
{\textbf{Square}}
&{147.0} / {190.8}
&{212.5} / {242.8}
&\textbf{388.1} / {401.4}
&{249.1} / {263.6}
&\textbf{56.3} / {89.5}\\
{\textbf{Circular}}
&\textbf{154.7} / \textbf{196.0}
&\textbf{216.8} / {248.9}
&{377.2} / {398.6}
&\textbf{251.2} / {264.5}
&{54.6} / {91.1}\\
{\textbf{Trapezoid}}
&{151.6} / {193.1}
&{215.2} / \textbf{255.8}
&{380.2} / \textbf{401.8}
&{250.0} / \textbf{266.1}
&{55.8} / \textbf{93.1}\\
\bottomrule
\end{tabular}
\end{center}
\vskip -3ex
\label{tab:shape}
\end{table}

\begin{table*}[!ht]
\setlength{\tabcolsep}{6mm}
\caption{Results across different Parts of the Shanghai Tech dataset. Higher MAE and MSE values indicate a stronger attack.}
\scriptsize
\centering
\subtable[Results on the Shanghai Tech Part A]{
\begin{tabular}{ccccccc}
\toprule
{\textbf{MAE / MSE}}&\multicolumn{6}{c}{\textbf{Target Model}}\\ 
\midrule
\textbf{Source Model} & MCNN                   & CSRNet & CAN & BL & DM-Count & SASNet\\ 
\midrule
\rowcolor{gray!20}
{Clean} 
& { 108.0} / {165.0 }         
& {  67.0} / {105.2 }           
& {  59.9} / {94.1  }           
& {  61.8} / {94.1  }             
& {  58.2} / {93.2  }            
& {  52.8} / {86.2  }           
\\
{CSRNet trained on Part A} 
& \textbf{147.0} / \textbf{190.8}   
& --
& \textbf{212.5} / \textbf{242.8} 
& \textbf{388.1} / \textbf{401.4}   
& \textbf{249.1} / \textbf{263.6}  
& \textbf{56.3} / \textbf{89.5} 
\\
{CSRNet trained on Part B} 
&{ 132.1} / {179.1 }   
&{ 155.0} / {174.3 } 
&{ 200.9} / {216.7 } 
&{ 264.7} / {279.4 }   
&{ 155.8} / {172.8 }  
&{ 53.4} / {86.9 }
\\ \bottomrule
\end{tabular}}

\subtable[Results on the Shanghai Tech Part B]{
\begin{tabular}{ccccccc}
\toprule
{\textbf{MAE / MSE}}&\multicolumn{6}{c}{\textbf{Target Model}}\\ 
\midrule
\textbf{Source Model} & MCNN                   & CSRNet & CAN & BL & DM-Count & SASNet\\ 
\midrule
\rowcolor{gray!20}
{Clean} 
& {28.3} / {38.7}         
& { 9.2} / {14.7}           
& { 7.5} / {11.9}           
& { 7.6} / {12.0}             
& { 7.3} / {11.8}            
& {6.4} / {9.9}           
\\
{CSRNet trained on Part A} 
& { 139.5} / {144.5 } 
& { 114.7} / {153.4 }  
& { 60.8} / {67.6 }   
& { 39.1} / {42.4 }     
& { 17.9} / {22.3 }    
& { 7.1} / \textbf{11.0 }   
\\
{CSRNet trained on Part B} 
& \textbf{162.8} / \textbf{167.9}   
& --
& \textbf{112.0} / \textbf{113.8} 
& \textbf{48.4} / \textbf{50.7}   
& \textbf{27.4} / \textbf{37.3}  
& \textbf{ 7.2} / {10.9}
\\ \bottomrule

\end{tabular}}
\vskip -3ex
\label{tab:across datasets}
\end{table*}

\begin{table*}[!ht]
\setlength{\tabcolsep}{7mm}
\caption{Results of attacks decreasing the crowd counting numbers on Shanghai Tech dataset Part A. The results on the diagonal are in white-box settings while the others are in black-box settings. Higher MAE and MSE values indicate a stronger attack.}
\scriptsize
\centering
\begin{tabular}{ccccccc}
\toprule
{\textbf{MAE / MSE}}&\multicolumn{6}{c}{\textbf{Target Model}}\\ 
\midrule
\textbf{{Source model}} & MCNN                   & CSRNet & CAN & BL & DM-Count & SASNet\\ 
\midrule
\rowcolor{gray!20}
{Clean} 
& {108.0} / {165.0}         
& { 67.0} / {105.2}           
& {59.9} / {94.1}           
& {61.8} / {94.1}             
& {58.2} / {93.2}            
& {52.8} / {86.2}           
\\ 
{MCNN}   
& \cellcolor{blue!20}{121.5} / \textbf{180.7} 
& { 69.7} / {109.6}  
& { 63.0} / {101.7}   
& {63.3} / {97.4}     
& {61.9} / {98.5}    
& {54.0} / {90.1}   
\\ 
{CSRNet} 
& {113.5} / {167.6}   
& \cellcolor{blue!20}\textbf{ 72.9} / \textbf{114.5} 
& { 65.4} / {105.6} 
& {64.2} / {98.1}   
& {62.4} / {98.5}  
& {55.1} / \textbf{92.6}   
\\ 
{CAN} 
& \textbf{122.1} / {172.3}   
& { 70.8} / {110.7} 
& \cellcolor{blue!20}\textbf{ 67.2} / \textbf{109.3} 
& \textbf{ 98.9} / \textbf{123.7}   
& \textbf{ 72.4} / \textbf{108.8}  
& {54.8} / {91.2}   
\\ 
{BL}              
& {109.5} / {167.6}  
& { 70.1} / {110.0}  
& { 63.6} / {102.3}   
& \cellcolor{blue!20}{64.7} / {99.2} 
& {62.5} / {99.4}  
& {54.7} / {91.1}   
\\ 
{DM-Count}  
& {110.2} / {168.5}   
& { 70.0} / {110.3}  
& { 63.8} / {102.7}   
& { 65.3} / {100.4}   
& \cellcolor{blue!20}{ 66.2} / {104.8} 
& {54.1} / {90.8}   
\\ 
{SASNet} 
& {110.5} / {168.1}   
& { 69.3} / {109.4}  
& {61.9} / {98.2}      
& {61.9} / {96.3}    
& {61.3} / {96.3}   
& \cellcolor{blue!20}\textbf{55.9} / {91.1} 
\\ \bottomrule

\end{tabular}

\label{tab:attacks-81-A}
\end{table*}

\paragraph{\textbf{Effectiveness across different datasets.}}
In practice, it is highly impossible that an adversary would have access to the training set used for training the target model. The utility of one attack will become poor if it does not generalize well across different datasets. We conduct a toy experiment using the various Parts of the Shanghai Tech dataset in order to evaluate the effectiveness of our adversarial patches under different data distributions. Using CSRNet, we first train our patches on one Part and then test them on the other. Note that Part A and Part B follow different distributions and the test models are only trained on data from one part.

\begin{figure}[!ht]
\begin{center}
\includegraphics[width=0.9\linewidth]{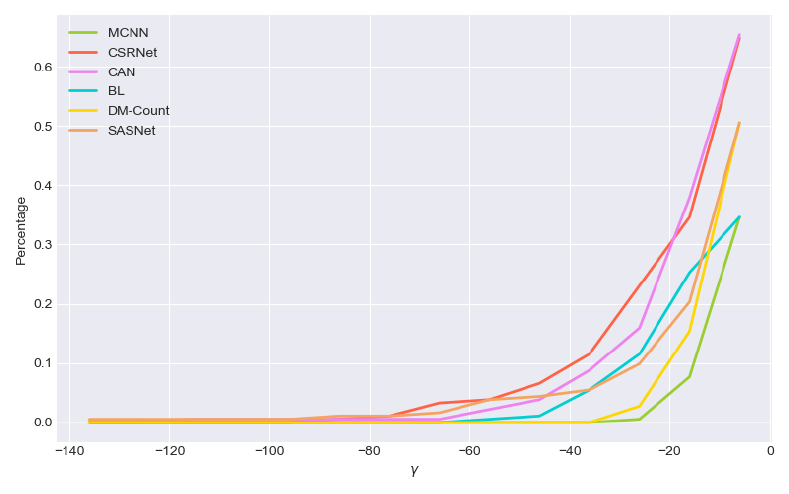}
\end{center} 
\vskip -3ex
\caption{Percentage of overestimation of crowd counting models under black-box PAP attack aiming to decrease the crowd numbers on Shanghai Tech Part A.}
\vskip -3ex
\label{fig:dec}
\end{figure}

As illustrated in Table \ref{tab:across datasets}, when tested across datasets, the performance of the black-box attack degrades slightly, however, it is still adequate to pose a threat to the clean scenes, demonstrating its potential threats in real-world applications.

\paragraph{\textbf{Decreasing crowd counting numbers with PAP}}

In the above section, we only report the results of attacks that increase the counting numbers. Besides, we can also generate adversarial patches that decrease the crowd counting numbers. Intuitively, in order to mislead the model to produce the zero-density response, we need to force the model to hardly recognize the crowd features and pay more attention to other non-people regions. Thus, we could simply modify the optimization direction as follows,
\begin{equation}
\label{eqn:total-loss2}
    \begin{split}
     \mathop{\arg\min}_{\delta} \mathcal{L}_{s} + \lambda\mathcal{L}_{p},
    \end{split}
\end{equation}
where $\mathcal L_s$ is the scale perception loss and $\mathcal{L}_{p}$ is the position perception loss. By minimizing the scale perception loss, we can generate our adversarial patches by weakening the model recognition for the crowds under different scales. As for minimizing the position perception loss, the crowd density guided model-shared attention will be suppressed while the attention towards the other objects will be enhanced. Therefore, based on the two loss items, our adversarial patches may also successfully lead the model to wrong estimations by decreasing the counting numbers.

To evaluate the effectiveness, we conduct experiments on the Shanghai Tech Part A. Except for the optimization direction, we followed all the settings in Section \ref{attack-setting}. Table \ref{tab:attacks-81-A} lists the attacking results. From the table, we found an interesting phenomenon that the black-box attacks may be stronger than the white-box for decreasing the predicted density (such as for BL and DM-Count, generating adversarial examples using CAN is more reliable). Further, we plot the overestimation percentage in Figure \ref{fig:dec}. Compared with Table \ref{tab:black-box attacks} and Figure \ref{fig:4}, adversarial patches aiming to decrease the counting numbers have relatively weaker attacking ability than those for increasing the density. We will study how to expand the adversarial impact in future work.

\section{Improving Crowd Counting with Perceptual Adversarial Patches}
\subsection{Overview}
Recent studies have revealed the fact that crowd counting models are still facing several challenges, including weak generalization abilities across datasets and robustness on complex backgrounds, which cast a shadow over the applications in practice \cite{gao2020cnn}. Some studies \cite{xie2020adversarial,chen2021robust} have shown that adversarial examples can also be used to improve model performance if harnessed in the right manner. Inspired by them, we aim to take the advantage of our perceptual adversarial patches and use them to improve the performance of crowd counting models. However, to improve image recognition and object detection models, current studies \cite{xie2020adversarial,chen2021robust} adopt multiple Batch Normalization (BN) branches to respectively handle clean and adversarial examples during adversarial training, which modifies the model architectures. They cannot be simply implemented in the crowd counting task, where most models do not have BN layers. Therefore, we adversarially train crowd counting models with our perceptual adversarial patches to improve the model performance without modifying architectures.

Specifically, we modify the standard adversarial training scheme \cite{madry2017towards} to adapt our PAP framework, which can be defined as follows:
\begin{equation}
\label{eqn:adv-train}
    \begin{split}
    \mathop{\min}_{\Theta}\mathbb{E}_{(\mathbf{x},\mathbf{I})\sim\mathbf{D}} \left\{ \mathop{\max}_{\delta}\mathcal{L}(f_{\Theta}(\mathbf x_{adv}),\mathbf{I})\right\},
    \end{split}
\end{equation}
where $\mathbf x_{adv}$ is the adversarial example (combined by $\mathbf x$ and $\delta$ via Eqn (\ref{eqn:xadv}), $\mathbf{I}$ is the ground truth density map, and ${\Theta}$ is the crowd counting model parameters. In the max manner, $\mathcal{L}$ refers to the loss for patch generation while it represents the loss for model optimization in the min manner. In practice, instead of solving the min-max optimization problem iteratively, we simply generate all the adversarial examples via the pre-trained model at the beginning, which could achieve better performance and take less time \footnote{More analyses can be found in appendix section E}.

Our perceptual adversarial patches can attack models under different crowd scale perceptions and disturb them to focus on the wrong position perception regions. Adversarial training with our patches is able to further enhance the model for the tolerance of perturbations brought from scales and positions. In other words, the enhanced crowd counting model with our adversarial patches could increase the perception generalization for multiple crowd scales and rectify their perceptions by better focusing on the crowd itself under noises. Therefore, it will better generalize to unseen scenarios with different crowd scales and pay more attention to crowd regions rather than complex backgrounds in natural scenes.

In the following sections, we aim to prove the effectiveness of our perceptual adversarial patches in benefiting the model performance. Specifically, we evaluate the generalization ability across datasets and robustness towards complex backgrounds of the enhanced crowd counting models. 

\begin{table}[!t]
\caption{MAE/MSE in cross-dataset evaluation. Lower MAE and MSE values indicate better generalization.}
\begin{center}
\setlength{\tabcolsep}{2mm}
\scriptsize
\subtable[Results of the models trained on Shanghai Tech Part A]{
\begin{tabular}{cccc}
\toprule
\centering\textbf{Method}&{\textbf{Shanghai Tech Part B}}&{\textbf{UCF-CC-50}}&{\textbf{Crowd Surveillance}}\\
\midrule
{Vanilla}
&22.8 / 34.3&417.7 / 664.2&24.9 / 52.2\\
{Cutout}
&18.0 / 27.9&396.7 / 615.0&19.5 / 40.0\\
{Cutmix}
&22.1 / 34.1&416.9 / 632.6&19.7 / 37.9\\
{Augmix}
&17.9 / 29.0&467.2 / 672.3&13.6 / 35.4\\
{PAT}
&23.7 / 35.1&443.0 / 625.5&30.6 / 69.1\\
{APAM-AT}
&23.6 / 35.1&421.9 / 664.9&25.0 / 53.2\\
\textbf{Ours}
&\textbf{17.5} / \textbf{27.5}&\textbf{382.0} / \textbf{594.9}&\textbf{12.7} / \textbf{30.9}\\
\bottomrule
\end{tabular}}

\subtable[Results of the models trained on Shanghai Tech Part B]{
\begin{tabular}{cccc}
\toprule
\centering\textbf{Method}&{\textbf{Shanghai Tech Part A}}&{\textbf{UCF-CC-50}}&{\textbf{Crowd Surveillance}}\\
\midrule
{Vanilla}
&142.4 / 241.3&1093.9 / 1405.6&11.2 / 22.7\\
{Cutout}
&153.1 / 272.3&1135.0 / 1454.5&11.5 / 23.9\\
{Cutmix}
&147.5 / 241.9&1129.4 / 1437.1&12.0 / 22.6\\
{Augmix}
&145.3 / 243.2&1060.2 / 1385.2&11.0 / 23.5\\
{PAT}
&145.7 / 249.6&1112.6 / 1445.9&13.5 / 25.8\\
{APAM-AT}
&142.6 / 243.4&1145.3 / 1459.6&13.1 / 24.8\\
\textbf{Ours}
&\textbf{129.8} / \textbf{220.5}&\textbf{717.9} / \textbf{1050.7}& \textbf{10.8} / \textbf{22.6}\\
\bottomrule
\end{tabular}}
\label{tab:gen}
\end{center}
\vskip -2ex
\end{table}

\paragraph{\textbf{Experimental settings.}} Following the settings in Section \ref{attack-setting}, we first generate adversarial patches on each image in the original training set and mix them to obtain the new training set (the ratio of adversarial examples and clean examples is 1:1). Then, we train the crowd counting model using the new training set. We select the multi-column based model MCNN and single-column based model DM-Count for evaluation and compare with two adversarial training methods: adversarial training with APAM generated adversarial patches (APAM-AT), PGD-$L_{\infty}$ adversarial training \cite{madry2017towards} (PAT, $iter=20,\alpha=0.002, \epsilon=8/255$), and three data augmentation methods: Cutout \cite{devries2017improved}, Cutmix \cite{yun2019cutmix}, and Augmix \cite{hendrycks2019augmix}. We faithfully follow the original settings for better implementation of the mentioned strategies. Moreover, we conduct the above methods on the same samples and use the same amount of extra data to train models for fair comparisons. Due to limited pages, we here only report the results of DM-Count \footnote{More results can be found in appendix section F}.

\begin{figure*}[!ht]
\begin{center}
\subfigure[Distractors]{
\includegraphics[width=0.15\linewidth]{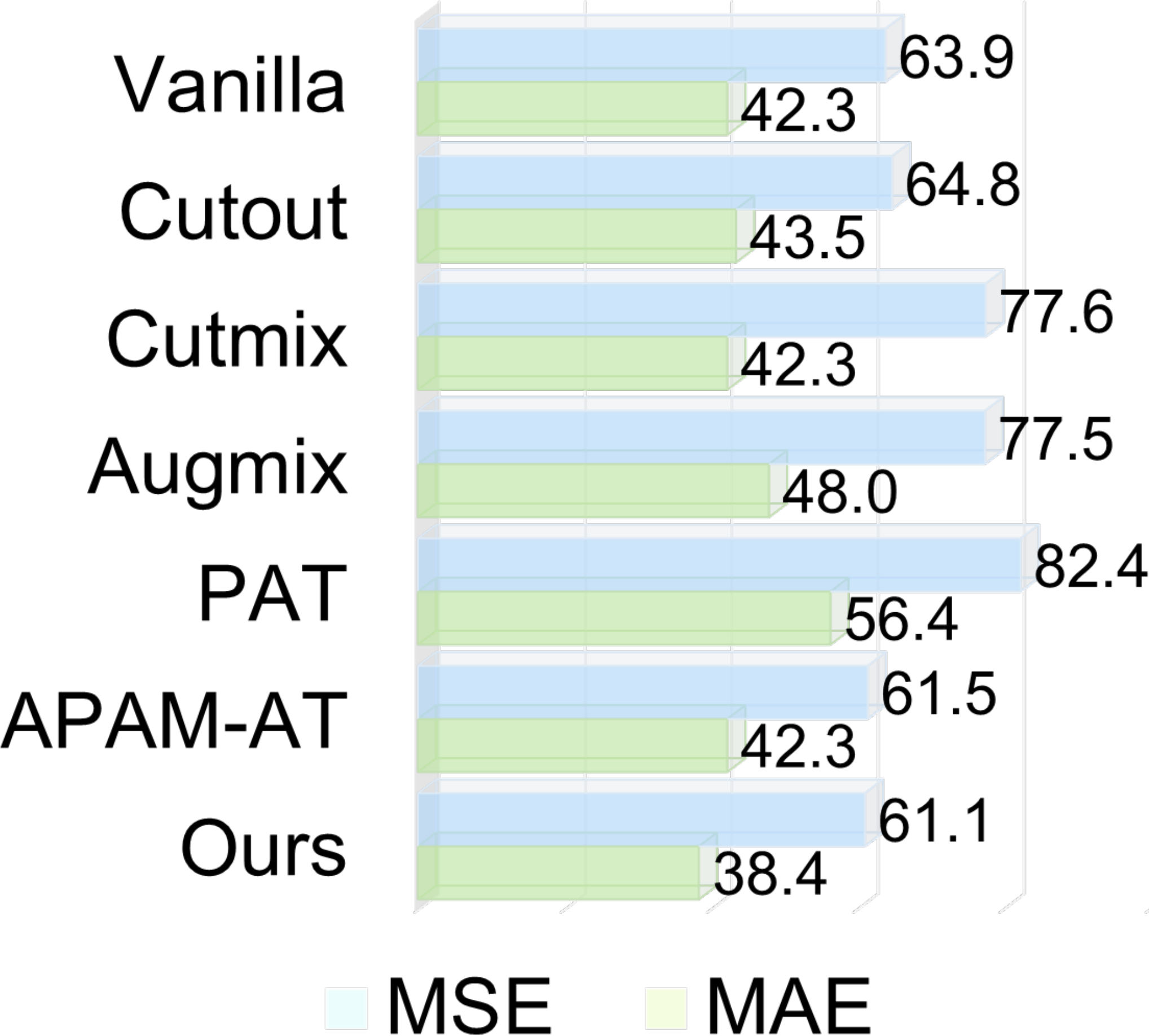}
}
\quad
\quad
\quad
\quad
\quad
\quad
\quad
\quad
\quad
\quad
\quad
\subfigure[Adverse Weather]{
\includegraphics[width=0.15\linewidth]{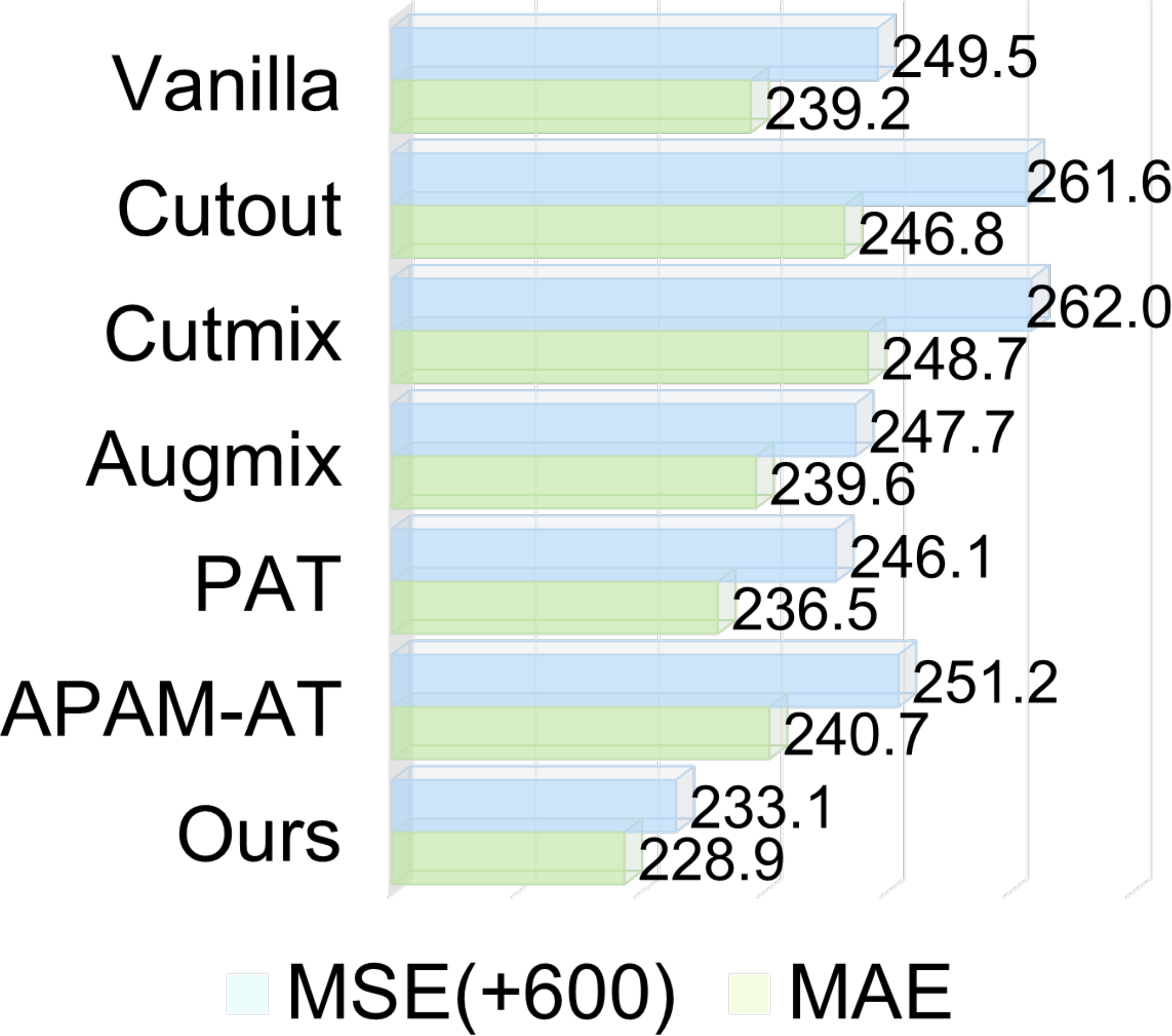}
}
\quad
\quad
\quad
\quad
\quad
\quad
\quad
\quad
\quad
\quad
\quad
\subfigure[Negative Samples]{
\includegraphics[width=0.15\linewidth]{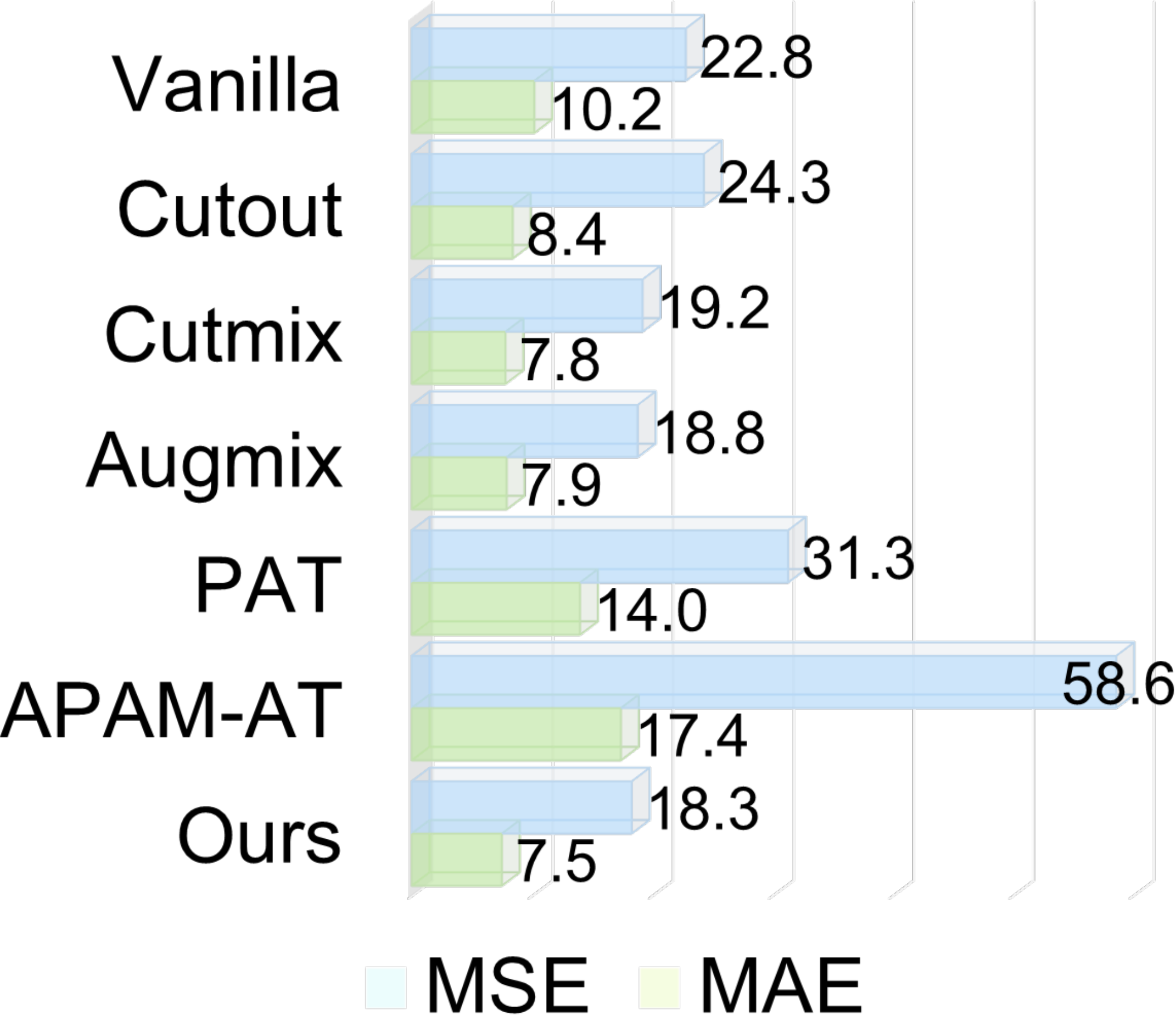}
}

\end{center}
\vskip -3ex
\caption{Model performance on images with complex backgrounds (\ie, distractors, adverse weathers, and negative samples). Lower MAE and MSE values indicate better robustness.}
\vskip -2ex
\label{fig:hard}
\end{figure*}

\begin{figure*}[!ht]
\begin{center}
\includegraphics[width=1.0\linewidth]{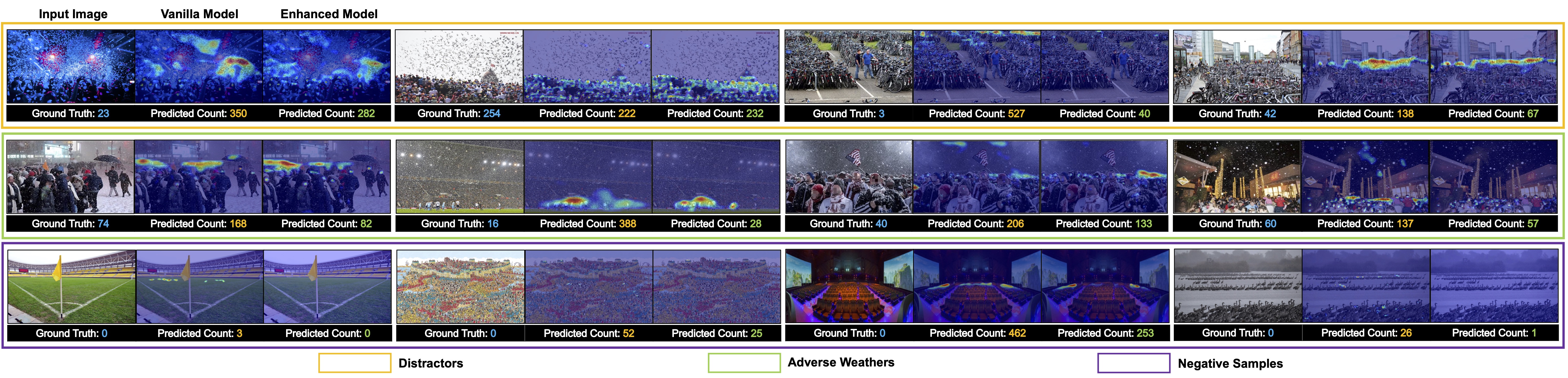}
\end{center} 
\vskip -2ex
\caption{The density map for models on the scenes with complex backgrounds. The model trained with our adversarial patch focus on the crowd more precisely, leading to better robustness.}
\vskip -2ex
\label{fig:hard2}
\end{figure*}

\subsection{Generalization across Datasets}

As images in different parts of the Shanghai Tech dataset were taken from different scenarios in different ways, we use Part A and Part B to conduct the cross-dataset performance evaluation. Besides, we also test the model performance on two other commonly-used crowd counting datasets, \ie, UCF-CC-50 \cite{idrees2013multi} and Crowd Surveillance \cite{yan2019perspective}.

As shown in Table \ref{tab:gen}, models trained with our PAP can significantly improve the generalization ability across datasets by large margins (at most \textbf{-376.0 MAE} and \textbf{-354.9 MSE}). We also outperform the adversarial training baselines (\eg, APAM-AT and PAT) which deteriorate the model generalization and the data augmentation techniques (\eg, Cutout, Cutmix, and Augmix).

\subsection{Robustness for Complex Backgrounds}
Following \cite{gao2020cnn}, we validate the model performance using three test sets including: 100 distractors and 191 adverse weather samples in JHU-CROWD++ \cite{sindagi2020jhu}, and 351 negative samples in NWPU \cite{wang2020nwpu}. Specifically, distractors are densely arranged other objects which may be confused for the crowd; adverse weather samples were taken under special weather conditions such as rain, snow, and haze; and negative samples do not contain any persons.

Figure \ref{fig:hard} shows the estimation errors on the mentioned three types of samples. Models trained with our PAP could improve robustness on all test sets (\textbf{-3.9 MAE} and \textbf{-2.8 MSE} on distractors, \textbf{-10.3 MAE} and \textbf{-16.4 MSE} on adverse weather samples, and \textbf{-2.7 MAE} and \textbf{-4.5 MSE} on negative samples), and also outperform other methods. Intuitively, adversarial training with our patches can help models to resist the crowd-like noises and focus on the real crowd patterns, resulting in stronger robustness on negative samples. As the visualization shown in Figure \ref{fig:hard2}, many non-target areas are highlighted on the density map of the vanilla model, such as the region of bicycles in the second row, whereas the enhanced model is significantly more robust to these distractors. The attention of the models trained with our PAP could focus on the human areas more accurately and thus the number of people predicted by the enhanced model is much closer to the ground truth. Therefore, our method can effectively facilitate the crowd counting model to perform better in the real world.

\section{Conclusion and Ethics Statement}
The premise of our work is that automated crowd counting is considered valuable, though, it may expose privacy risks in the application and data collection process. We oppose its application for malicious surveillance; instead, we focus on scenarios that are beneficial to humans. For instance, traffic monitoring and public safety are significantly meaningful use-cases for automated crowd counting. This technology can be utilized for crowd flow monitoring which promises better traffic planning for automated driving systems. Besides, effective crowd counting can prevent stampede accidents caused by excessive crowd density. Therefore, we believe that automated crowd counting is developed for the better. In addition, we have made every effort to protect personal information during the collection and use of data. All images used in this paper are from public datasets or collected legally. We seek the consent of the subject when capturing images in the physical world and mask the face information in our paper for privacy protection.

Adversarial attacks, as an effective way to discover security vulnerabilities in practice, will facilitate researchers to pay more attention to the robustness of the models. Given this, to generate strong transferable attacks for crowd counting models, this paper proposes the Perceptual Adversarial Patch (PAP) generation framework to learn the model-invariant features by exploiting both the model scale perception and position perception. To validate the effectiveness of our proposed method, we conduct extensive experiments in both the digital and physical world, which shows that PAP achieves state-of-the-art performance. Through our attack, it is demonstrated that existing adversarial defense strategies on the regression task are not infallible. Moreover, it is worthwhile to further explore how to effectively define robustness metrics on the crowd counting task. We believe this paper will inspire future research on these aspects.

Additionally, our attack can be utilized to protect privacy. We can successfully disrupt malicious surveillance systems to protect crowd information, thus preventing the infringement of the right to a public meeting. In this case, we still have a lot of obstacles to overcome, such as how to make our adversarial patch appear natural enough to go undetected and how to efficiently cover all observation locations. Even while our work suggests putting patches on clothing or posters, we anticipate more effective methods of patch generation to prevent anomalous warnings brought on by human perception.

From another perspective, we expect that the automated crowd counting models have a strong generalization to cope with changing real-world conditions. However, existing deep learning methods do suffer from overfitting known data distributions. Nevertheless, the deep crowd counting approach based on density map estimation has been the most accurate and efficient way. In contrast to most previous studies, we surprisingly find that adversarial training with our patches can benefit model performance, revealing another approach to exploit adversarial attack techniques for social positive. We utilize our adversarial patches as beneficial enhancement data to enhance the model generalization to unknown crowd scales and robustness towards complex backgrounds, leading to better application in reality. Though providing a preliminary explanation, we are interested in investigating the nature and mechanism of the observation, and we leave it as future work.

\begin{acks}
This work was supported by The National Key Research and Development Plan of China (2020AAA0103502), and National Natural Science Foundation of China (62022009 and 61872021).
\end{acks}

\clearpage
\bibliographystyle{ACM-Reference-Format}
\bibliography{acmart}

\clearpage
\appendix

\section{Attacks with different patch sizes}

For the limited page sizes, we only show the results of our PAP with the size of $81\times81$ in the main body of the paper. Here, we provide the results on Shanghai Tech Part A with different patch sizes ($40\times40$, $163\times163$) following \cite{wu2021towards} in the digital world. As illustrated in Table \ref{tab:attacks-40-A} and Table \ref{tab:attacks-163-A}, our adversarial patches have strong attacking ability under both white-box and black-box settings, which outperforms APAM \cite{wu2021towards} by large margins. 

\begin{table*}[!ht]
\setlength{\tabcolsep}{5mm}
\caption{Results of attacks on Shanghai Tech Part A with $40\times40$ patch size. The results on the diagonal are in white-box settings while the others are in black-box settings. Higher MAE and MSE values indicate a stronger attack.}
\scriptsize
\centering
\begin{tabular}{cccccccc}
\toprule
\multicolumn{2}{c}{\textbf{MAE / MSE}}&\multicolumn{6}{c}{\textbf{Target Model}}\\ 
\midrule
\textbf{{Source model}} & \textbf{Method} & MCNN                   & CSRNet & CAN & BL & DM-Count & SASNet\\ 
\midrule
\rowcolor{gray!20}
\multicolumn{2}{c}{Clean} 
& { 108.0} / {165.0 }         
& {  67.0} / {105.2 }           
& {  59.9} / {94.1  }           
& {  61.8} / {94.1  }             
& {  58.2} / {93.2  }            
& {  52.8} / {86.2  }           
\\
\multirow{2}{*}{MCNN} 
& APAM   
& \cellcolor{blue!20}{ 111.5} / {170.5 }            
& {  69.5} / {109.7 }           
& {  60.9} / {96.6  }
& {  62.5} / {95.6  }             
& {  60.3} / {96.2  }           
& {  53.6} / {87.8  }           
\\ \multirow{2}{*}{} 
& \textbf{Ours}   
& \cellcolor{blue!20}\textbf{ 201.0} / \textbf{245.9 } 
& \textbf{ 142.8} / \textbf{230.1 }  
& \textbf{ 141.3} / \textbf{230.7 }   
& \textbf{ 135.5} / \textbf{216.6 }     
& \textbf{ 142.5} / \textbf{224.6 }    
& \textbf{ 110.2} / \textbf{195.9 }   
\\
\multirow{2}{*}{CSRNet} 
& APAM   
& { 109.6} / {168.0 }            
& \cellcolor{blue!20}{  69.3} / {109.3 }           
& {  61.0} / {96.5  }           
& {  62.4} / {95.5  }             
& {  60.1} / {96.2  }            
& {  53.6} / {87.5  }          
\\ \multirow{2}{*}{} 
& \textbf{Ours}   
& \textbf{ 118.6} / \textbf{182.1 }   
& \cellcolor{blue!20}\textbf{ 116.6} / \textbf{192.6 } 
& \textbf{ 122.4} / \textbf{211.7 } 
& \textbf{ 107.8} / \textbf{178.3 }   
& \textbf{ 110.5} / \textbf{189.7 }  
& \textbf{ 107.3} / \textbf{193.6 }   
\\ 
\multirow{2}{*}{CAN} 
& APAM   
& { 109.5} / {167.0 }            
& {  69.1} / {108.9 }           
& \cellcolor{blue!20}{  61.2} / {98.2  }           
& {  64.1} / {96.4  }             
& {  60.0} / {96.0  }            
& {  53.5} / {87.7  }           
\\ \multirow{2}{*}{} 
& \textbf{Ours}   
& \textbf{ 118.4} / \textbf{181.7 }   
& \textbf{ 111.3} / \textbf{195.6 } 
&\cellcolor{blue!20} \textbf{ 116.9} / \textbf{199.0 } 
& \textbf{ 108.9} / \textbf{181.6 }   
& \textbf{ 114.3} / \textbf{196.3 }  
& \textbf{ 107.0} / \textbf{193.4 }   
\\ 
\multirow{2}{*}{BL}              
& APAM   
& { 109.6} / {167.6 }             
& {  69.2} / {109.3 }           
& {  60.9} / {96.3  }           
& \cellcolor{blue!20}{  63.1} / {96.4  }             
& {  60.2} / {97.3  }            
& {  53.6} / {87.8  }  
\\ \multirow{2}{*}{} 
& \textbf{Ours}   
& \textbf{ 118.8} / \textbf{182.2 }  
& \textbf{ 134.2} / \textbf{222.5 }  
& \textbf{ 139.0} / \textbf{228.4 }   
& \cellcolor{blue!20}\textbf{ 195.7} / \textbf{224.8 } 
& \textbf{ 125.3} / \textbf{179.6 }  
& \textbf{ 109.3} / \textbf{195.3 }   
\\ 
\multirow{2}{*}{DM-Count}  
& APAM   
& { 109.4} / {167.8 }            
& {  69.3} / {109.5 }           
& {  61.0} / {96.6  }           
& {  62.9} / {95.6  }             
& \cellcolor{blue!20}{  62.4} / {98.0  }            
& {  53.5} / {87.8  }           
\\ \multirow{2}{*}{} 
& \textbf{Ours}   
& \textbf{ 119.0} / \textbf{182.4 }   
& \textbf{ 135.3} / \textbf{224.3 }  
& \textbf{ 138.9} / \textbf{228.5 }   
& \textbf{ 148.4} / \textbf{189.2 }   
& \cellcolor{blue!20}\textbf{ 144.5} / \textbf{186.6 } 
& \textbf{ 109.2} / \textbf{195.2 }   
\\ 
\multirow{2}{*}{SASNet} 
& APAM   
& { 109.9} / {183.2 }            
& {  69.2} / {108.9 }           
& {  61.1} / {96.1  }  
& {  63.2} / {96.3  }             
& {  62.3} / {98.3  }           
& \cellcolor{blue!20}{  53.7} / {87.6  }           
\\ \multirow{2}{*}{} 
& \textbf{Ours}   
& \textbf{ 119.3} / \textbf{183.2 }   
& \textbf{ 142.1} / \textbf{229.4 }  
& \textbf{ 141.4} / \textbf{230.8 }      
& \textbf{ 133.6} / \textbf{214.7 }    
& \textbf{ 142.0} / \textbf{224.3 }   
& \cellcolor{blue!20}\textbf{ 106.4} / \textbf{192.5 } 
\\ \bottomrule

\end{tabular}

\label{tab:attacks-40-A}
\end{table*}

\begin{table*}[!ht]
\setlength{\tabcolsep}{5mm}
\caption{Results of attacks on Shanghai Tech Part A with $163\times163$ patch size. The results on the diagonal are in white-box settings while the others are in black-box settings. Higher MAE and MSE values indicate a stronger attack.}
\scriptsize
\centering
\begin{tabular}{cccccccc}
\toprule
\multicolumn{2}{c}{\textbf{MAE / MSE}}&\multicolumn{6}{c}{\textbf{Target Model}}\\ 
\midrule
\textbf{{Source model}} & \textbf{Method} & MCNN                   & CSRNet & CAN & BL & DM-Count & SASNet\\ 
\midrule
\rowcolor{gray!20}
\multicolumn{2}{c}{Clean} 
& {108.0} / {165.0}         
& { 67.0} / {105.2}           
& {59.9} / {94.1}           
& {61.8} / {94.1}             
& {58.2} / {93.2}            
& {52.8} / {86.2}           
\\
\multirow{2}{*}{MCNN} 
& APAM   
& \cellcolor{blue!20}{432.1} / {495.3}            
& { 80.1} / {126.9}           
& { 76.2} / {121.6}
& { 69.7} / {108.2}             
& { 70.6} / {108.4}           
& { 63.2} / {103.4}           
\\ \multirow{2}{*}{} 
& \textbf{Ours}   
& \cellcolor{blue!20}\textbf{4431.8} / \textbf{4470.2} 
& \textbf{ 80.8} / \textbf{127.7}  
& \textbf{ 76.9} / \textbf{123.9}   
& \textbf{ 70.5} / \textbf{108.4}     
& \textbf{ 71.2} / \textbf{109.3}    
& \textbf{ 63.3} / \textbf{104.9}   
\\ 
\multirow{2}{*}{CSRNet} 
& APAM   
& {354.6} / {379.1}            
& \cellcolor{blue!20}{463.5} / {589.4}           
& {127.4} / {158.1}           
& {613.3} / {626.7}             
& {368.5} / {384.0}            
& { 61.8} / {100.8}          
\\ \multirow{2}{*}{} 
& \textbf{Ours}   
& \textbf{463.8} / \textbf{484.2}   
& \cellcolor{blue!20}\textbf{2234.4} / \textbf{2268.4} 
& \textbf{660.5} / \textbf{736.2} 
& \textbf{1246.8} / \textbf{1255.0}   
& \textbf{626.9} / \textbf{639.3}  
& \textbf{63.4} / \textbf{101.9}   
\\ 
\multirow{2}{*}{CAN} 
& APAM   
& {441.5} / {464.4}            
& {330.6} / {346.8}           
& \cellcolor{blue!20}{410.2} / {528.4}           
& {719.4} / {728.0}             
& {439.6} / {453.0}            
& {63.7} / {99.5}           
\\ \multirow{2}{*}{} 
& \textbf{Ours}   
& \textbf{530.3} / \textbf{551.3}   
& \textbf{1070.1} / \textbf{1083.2} 
& \cellcolor{blue!20}\textbf{2084.0} / \textbf{2142.0} 
& \textbf{1051.3} / \textbf{1058.9}   
& \textbf{472.3} / \textbf{485.5}  
& \textbf{ 71.9} / \textbf{105.2}   
\\
\multirow{2}{*}{BL}              
& APAM   
& {230.2} / {262.4}             
& { 77.9} / {116.8}           
& { 72.7} / {111.2}           
& \cellcolor{blue!20}{290.4} / {313.1}             
& {370.0} / {387.0}            
& { 61.9} / {100.2}  
\\ \multirow{2}{*}{} 
& \textbf{Ours}   
& \textbf{285.7} / \textbf{311.3}  
& \textbf{238.0} / \textbf{255.0}  
& \textbf{213.0} / \textbf{261.5}   
& \cellcolor{blue!20}\textbf{5213.4} / \textbf{5300.0} 
& \textbf{1987.5} / \textbf{2003.3}  
& \textbf{ 62.0} / \textbf{103.2}   
\\ 
\multirow{2}{*}{DM-Count}  
& APAM   
& {204.3} / {238.5}            
& {67.2} / {111.1}           
& { 71.1} / {113.4}           
& {114.5} / {140.8}             
& \cellcolor{blue!20}{207.4} / {236.5}            
& { 63.9} / {100.1}           
\\ \multirow{2}{*}{} 
& \textbf{Ours}   
& \textbf{249.7} / \textbf{278.3}   
& \textbf{266.9} / \textbf{284.5}  
& \textbf{207.1} / \textbf{262.0}   
& \textbf{2740.0} / \textbf{2767.2}   
& \cellcolor{blue!20}\textbf{3168.0} / \textbf{3194.2} 
& \textbf{ 73.2} / \textbf{103.8}   
\\ 
\multirow{2}{*}{SASNet} 
& APAM   
& {111.4} / {171.6}            
& { 78.9} / {127.6}           
& { 74.2} / {118.9}  
& { 68.7} / {107.2}             
& { 71.1} / {111.4}           
& \cellcolor{blue!20}{ 71.8} / {119.3}           
\\ \multirow{2}{*}{} 
& \textbf{Ours}   
& \textbf{444.4} / \textbf{463.7}   
& \textbf{388.2} / \textbf{402.4}  
& \textbf{142.7} / \textbf{167.7}      
& \textbf{727.8} / \textbf{744.0}    
& \textbf{423.0} / \textbf{438.1}   
& \cellcolor{blue!20}\textbf{1801.9} / \textbf{1813.9} 
\\ \bottomrule
\end{tabular}

\label{tab:attacks-163-A}
\end{table*}

\section{Attacks against defended models}
To better evaluate our PAP performance, we conduct attacks under different defensive methods. Due to limited computing resources, we select five models listed in the main paper, \ie, MCNN, CSRNet, CAN, BL and DM-Count, as the target models. First, we choose the adversarial training scheme \cite{madry2017towards}, which has been proved to be the best empirical defense. In the adversarial patch scene, we iteratively generate white-box adversarial patches during the model training loop to improve the robustness. Referring \cite{wu2021towards}, the training loss can be formulated as follows,
\begin{equation}
    \mathcal L = \lambda \cdot \mathcal L_{clean} + (1 - \lambda) \cdot \mathcal L_{adv}
\label{eqn:1}
\end{equation}
For the total epochs $E$, we set $\lambda=1$ in $[0, 0.25E]$ to warm up, and then slowly decrease $\lambda$ from 1 to 0.5 in $[0.25E, 0.5E]$, finally, maintain $\lambda=0.5$ to finish the remaining epochs. We set $\alpha = 0.01$ and $T=5$ for our PAP attack. Except as mentioned above, the training of all models follows the original setup.

As for the certified method, several studies \cite{levine2020randomized,mccoyd2020minority,xiang2021patchcleanser} have been devoted to improving the robustness against adversarial patches in the classification task. However, these approaches have strict constraints on the patch size and require classification-oriented information, \eg, output probabilities and category labels. This leads to the fact that they cannot be directly applied to the regression scenario like crowd counting. Therefore, we consider the randomized ablation method, a general certified defense method for crowd counting models proposed in \cite{wu2021towards} as our target. We follow \cite{wu2021towards} re-training the models with the hyperparameter $k=45$.

Table \ref{tab:3} and \ref{tab:4} list the results for the adversarial training and randomized ablation, respectively. It can be seen that our PAP still performs strong attacking ability, which will facilitate research into better defenses.

\begin{table*}[!ht]
\setlength{\tabcolsep}{9mm}
\caption{Results of attacks towards adversarial training on Shanghai Tech Part A. The results on the diagonal are in white-box settings while the others are in black-box settings. Higher MAE and MSE values indicate a stronger attack.}
\scriptsize
\centering
\begin{tabular}{cccccc}
\toprule
{\textbf{MAE / MSE}}&\multicolumn{5}{c}{\textbf{Target Model}}\\ 
\midrule
\textbf{{Source model}} & MCNN                   & CSRNet & CAN & BL & DM-Count\\ 
\midrule
\rowcolor{gray!20}
{Clean} 
& {112.7} / {169.6}         
& { 71.0} / {109.7}           
& { 69.4} / {108.7}           
& { 67.2} / {109.1}             
& { 73.0} / {117.8}            
\\
{MCNN} 
& \cellcolor{blue!20}{1580.9} /{1781.1} 
& { 72.1} / {112.7}  
& { 70.8} / {110.5}   
& { 67.7} / {109.2}     
& { 74.1} / {120.1}    
\\
{CSRNet} 
& {171.2} / {215.7}   
& \cellcolor{blue!20}{535.7} / {576.5} 
& {263.7} / {292.3} 
& {372.6} / {387.6}   
& {102.2} / {140.5}  
\\ 
{CAN} 
& {168.5} / {214.1}   
& {318.9} / {340.7} 
&\cellcolor{blue!20} {427.5} / {478.1} 
& {319.3} / {335.0}   
& { 88.6} / {130.6}  
\\ 
{BL}              
& {130.1} / {182.5}  
& {120.8} / {146.1}  
& {144.7} / {168.9}   
& \cellcolor{blue!20}{1075.9} / {1103.3} 
& {122.6} / {154.5}  
\\ 
{DM-Count}  
& {116.9} / {172.0}   
& { 89.9} / {122.2}  
& { 87.4} / {119.4}   
& {230.1} / {257.8}   
& \cellcolor{blue!20}{257.8} / {285.1} 
\\ \bottomrule

\end{tabular}

\label{tab:3}
\end{table*}

\begin{table*}[!ht]
\setlength{\tabcolsep}{9mm}
\caption{Results of attacks towards randomized ablation on Shanghai Tech Part A. The results on the diagonal are in white-box settings while the others are in black-box settings. Higher MAE and MSE values indicate a stronger attack.}
\scriptsize
\centering
\begin{tabular}{cccccc}
\toprule
{\textbf{MAE / MSE}}&\multicolumn{5}{c}{\textbf{Target Model}}\\ 
\midrule
\textbf{{Source model}} & MCNN                   & CSRNet & CAN & BL & DM-Count\\ 
\midrule
\rowcolor{gray!20}
{Clean} 
& {117.3} / {185.4}         
& { 75.3} / {134.9}           
& { 74.2} / {117.8}           
& { 73.7} / {108.1}             
& { 68.9} / {105.5}            
\\
{MCNN} 
& \cellcolor{blue!20}{249.1} /{378.3} 
& {148.8} / {214.1}  
& {167.1} / {258.0}   
& {218.4} / {323.9}     
& {166.0} / {255.3}    
\\
{CSRNet} 
& {247.3} / {376.5}   
& \cellcolor{blue!20}{621.3} / {677.7} 
& {338.4} / {472.1}
& {228.4} / {321.7}   
& {180.9} / {264.3}  
\\ 
{CAN} 
& {249.0} / {378.1}   
& {134.0} / {176.5} 
&\cellcolor{blue!20} {699.6} / {760.5} 
& {225.8} / {322.7}   
& {185.0} / {274.3}  
\\ 
{BL}              
& {248.8} / {378.9}  
& {181.6} / {268.3}  
& {361.1} / {508.1}   
& \cellcolor{blue!20}{328.4} / {422.0} 
& {183.3} / {275.5}  
\\ 
{DM-Count}  
& {259.0} / {388.2}   
& {169.0} / {254.2}  
& {365.5} / {510.4}   
& {222.9} / {321.7}   
& \cellcolor{blue!20}{386.5} / {525.5} 
\\ \bottomrule

\end{tabular}

\label{tab:4}
\end{table*}

\begin{table*}[!ht]
\scriptsize
\caption{Results of different transferable attacks on Shanghai Tech dataset Part A. The results on the diagonal are in white-box settings while the others are in black-box settings. Higher MAE and MSE values indicate a stronger attack.}
\centering
\setlength{\tabcolsep}{5mm}
\begin{tabular}{cccccccc}
\toprule
\multicolumn{2}{c}{\textbf{MAE / MSE}}&\multicolumn{6}{c}{\textbf{Target Model}}\\ 
\midrule
\textbf{{Source model}} & \textbf{Method} & MCNN                   & CSRNet & CAN & BL & DM-Count & SASNet\\
\midrule
\rowcolor{gray!20}
\multicolumn{2}{c}{Clean}
&{108.0} / {165.0}         
&{ 67.0} / {105.2}           
&{59.9} / {94.1}           
&{61.8} / {94.1}             
&{58.2} / {93.2}            
&{52.8} / {86.2}           
\\\midrule
\multirow{5}{*}{MCNN} 
& MIGM
&\cellcolor{blue!20}{702.5} / {728.8}
&{69.8} / {109.3}
&{62.2} / {100.9}
&{63.4} / {96.7}
&{60.8} / {95.9}
&{53.5} / {89.3}
\\ \multirow{5}{*}{} 
& NIGM
&\cellcolor{blue!20}{673.0} / {701.9}
&\textbf{69.8} / \textbf{109.5}
&{62.2} / {100.8}
&{62.9} / {97.0}
&{61.3} / {95.9}
&{54.3} / {89.0}
\\ \multirow{5}{*}{} 
& TI-NIGM
&\cellcolor{blue!20}{182.4} / {220.5}
&{69.5} / {109.0}
&{62.7} / {100.9}
&{63.2} / \textbf{97.2}
&{61.0} / {96.8}
&{53.9} / {89.5}
\\ \multirow{5}{*}{} 
& NAA
&\cellcolor{blue!20}{700.5} / {731.3}
&{69.8} / {109.4}
&{62.0} / {100.6}
&{62.9} / {97.0}
&{60.2} / {95.8}
&{53.5} / {89.2}
\\ \multirow{5}{*}{} 
& \textbf{Ours}   
& \cellcolor{blue!20}\textbf{908.7} / \textbf{989.1} 
& { 69.6} / {109.3}  
& \textbf{ 62.9} / \textbf{101.3}   
& \textbf{64.1} / {96.9}     
& \textbf{62.0} / \textbf{96.8}    
& \textbf{54.3} / \textbf{89.9}   
\\\midrule
\multirow{5}{*}{CSRNet} 
& MIGM
&{118.5} / {170.3}
&\cellcolor{blue!20}{305.5} / {326.7}
&{104.2} / {128.8}
&{239.8} / {253.1}
&{185.4} / {201.2}
&{54.2} / {89.9}
\\ \multirow{5}{*}{} 
& NIGM
&{125.6} / {175.2}
&\cellcolor{blue!20}{312.3} / {333.4}
&{87.4} / {110.7}
&{267.9} / {281.2}
&{204.4} / {220.4}
&{54.2} / \textbf{90.0}
\\ \multirow{5}{*}{} 
& TI-NIGM
&{110.3} / {167.7}
&\cellcolor{blue!20}{107.3} / {133.5}
&{60.7} / {95.2}
&{75.8} / {104.3}
&{72.7} / {103.2}
&{53.8} / {88.7}
\\ \multirow{5}{*}{} 
& NAA
&{119.4} / {171.1}
&\cellcolor{blue!20}{298.6} / {320.7}
&{93.8} / {120.4}
&{201.5} / {222.4}
&{165.1} / {184.2}
&{53.9} / {89.5}
\\ \multirow{5}{*}{} 
& \textbf{Ours}   
& \textbf{147.0} / \textbf{190.8}   
& \cellcolor{blue!20}\textbf{568.4} / \textbf{613.8} 
& \textbf{212.5} / \textbf{242.8} 
& \textbf{388.1} / \textbf{401.4}   
& \textbf{249.1} / \textbf{263.6}  
& \textbf{56.3} / {89.5}   
\\\midrule
\multirow{5}{*}{CAN} 
& MIGM
&{145.2} / {188.9}
&{255.6} / {273.9}
&\cellcolor{blue!20}{426.0} / {453.3}
&{385.9} / {397.7}
&{203.8} / {218.3}
&{54.2} / {88.5}
\\ \multirow{5}{*}{} 
& NIGM
&{146.9} / {190.2}
&{267.2} / {285.3}
&\cellcolor{blue!20}{438.2} / {466.2}
&{395.8} / {407.7}
&{204.0} / {218.3}
&{53.8} / {88.3}
\\ \multirow{5}{*}{} 
& TI-NIGM
&{111.0} / {169.7}
&{69.5} / {109.1}
&\cellcolor{blue!20}{64.2} / {103.4}
&{63.3} / {97.2}\
&{61.6} / {97.6}
&{54.5} / {88.5}
\\ \multirow{5}{*}{} 
& NAA
&{139.3} / {184.5}
&{257.3} / {276.9}
&\cellcolor{blue!20}{417.6} / {445.6}
&{398.8} / {411.8}
&\textbf{220.5} / \textbf{234.8}
&{54.3} / {88.0}
\\ \multirow{5}{*}{} 
& \textbf{Ours}   
& \textbf{147.6} / \textbf{191.5}   
& \textbf{321.0} / \textbf{341.7} 
& \cellcolor{blue!20}\textbf{513.3} / \textbf{545.3} 
& \textbf{412.7} / \textbf{424.8}   
& {218.8} / {233.9}  
& \textbf{56.2} / \textbf{88.8}   
\\\midrule
\multirow{5}{*}{BL}              
& MIGM
&\textbf{120.4} / \textbf{171.6}
&{77.9} / {111.0}
&{72.2} / {105.6}
&\cellcolor{blue!20}{1001.6} / {1023.7}
&{491.9} / {504.9}
&{54.5} / {89.4}
\\ \multirow{5}{*}{} 
& NIGM
&{119.1} / {170.9}
&{74.2} / {107.6}
&{68.1} / {102.3}
&\cellcolor{blue!20}{1023.9} / {1051.3}
&{511.7} / {526.5}
&{53.7} / {89.5}
\\ \multirow{5}{*}{} 
& TI-NIGM
&{109.9} / {167.8}
&{74.5} / {109.8}
&{65.4} / {98.1}
&\cellcolor{blue!20}{111.1} / {130.6}
&{77.9} / {107.3}
&{54.5} / {89.9}
\\ \multirow{5}{*}{} 
& NAA
&{121.0} / {172.6}
&{78.5} / {110.2}
&{69.3} / {103.4}
&\cellcolor{blue!20}{1022.6} / {1044.9}
&{508.9} / {522.1}
&{53.7} / {89.6}
\\ \multirow{5}{*}{} 
& \textbf{Ours}   
& {119.0} / {170.7}  
& \textbf{ 79.6} / \textbf{111.5}  
& \textbf{ 73.1} / \textbf{106.6}   
& \cellcolor{blue!20}\textbf{1090.9} / \textbf{1171.6} 
& \textbf{519.7} / \textbf{541.6}  
& \textbf{54.5} / \textbf{90.0}   
\\\midrule
\multirow{5}{*}{DM-Count}  
& MIGM
&{116.2} / {169.4}
&{70.9} / {106.2}
&{65.8} / {100.8}
&{703.9} / {713.2}
&\cellcolor{blue!20}{742.0} / {779.1}
&{55.9} / {90.3}
\\ \multirow{5}{*}{} 
& NIGM
&\textbf{117.8} / \textbf{170.0}
&{70.3} / {106.5}
&{64.8} / {100.1}
&{701.0} / {731.2}
&\cellcolor{blue!20}{744.7} / {761.9}
&{55.8} / {90.0}
\\ \multirow{5}{*}{} 
& TI-NIGM
&{111.2} / {169.3}
&{68.9} / {108.7}
&{62.3} / {99.9}
&{63.3} / {96.6}
&\cellcolor{blue!20}{62.4} / {98.1}
&{53.9} / {89.5}
\\ \multirow{5}{*}{} 
& NAA
&{117.0} / {170.1}
&{70.6} / {106.5}
&{65.3} / {101.1}
&{737.7} / {746.0}
&\cellcolor{blue!20}{744.0} / {781.4}
&{56.4} / {90.8}
\\ \multirow{5}{*}{} 
& \textbf{Ours}   
& {115.6} / {169.6}   
& \textbf{ 87.6} / \textbf{119.3}  
& \textbf{ 82.8} / \textbf{115.0}   
& \textbf{747.5} / \textbf{793.6}   
& \cellcolor{blue!20}\textbf{751.3} / \textbf{784.9} 
& \textbf{56.5} / \textbf{90.8}   
\\\midrule
\multirow{5}{*}{SASNet} 
& MIGM
&{123.2} / {173.2}
&\textbf{71.5} / {108.3}
&{61.7} / {95.7}
&{65.5} / {95.3}
&{69.0} / {91.3}
&\cellcolor{blue!20}{122.5} / {147.5}
\\ \multirow{5}{*}{} 
& NIGM
&\textbf{124.6} / \textbf{174.2}
&{71.0} / {106.5}
&{60.5} / {93.9}
&{63.9} / {93.5}
&{72.0} / {92.1}
&\cellcolor{blue!20}{132.0} / {147.0}
\\ \multirow{5}{*}{} 
& TI-NIGM
&{110.4} / {168.7}
&{69.4} / {108.2}
&{61.9} / {99.5}
&{64.1} / {97.5}
&{61.0} / {97.6}
&\cellcolor{blue!20}{56.1} / {89.2}
\\ \multirow{5}{*}{} 
& NAA
&{110.6} / {169.3}
&{69.9} / \textbf{109.8}
&{61.0} / {99.5}
&{63.5} / {97.4}
&{62.1} / {98.1}
&\cellcolor{blue!20}{53.8} / {90.1}
\\ \multirow{5}{*}{} 
& \textbf{Ours}   
& {112.8} / {169.4}   
& { 69.5} / {109.0}  
& \textbf{ 62.6} / \textbf{100.3}      
& \textbf{69.9} / \textbf{99.8}    
& \textbf{ 75.1} / \textbf{105.4}   
& \cellcolor{blue!20}\textbf{200.0} / \textbf{220.3} 
\\ \bottomrule
\end{tabular}
\label{tab:5}
\end{table*}

\section{Comparing with other transferable attacks}
In the main paper, we conduct experiments to compare our PAP with six other methods designed for transferable attacks (MIGM \cite{dong2018boosting}, NIGM \cite{lin2019nesterov}, TI-NIGM \cite{dong2019evading}, NAA \cite{zhang2022improving}, Avg-Dens, and MGAA \cite{yuan2021meta}). For a fair comparison, all hyperparameters are set the same as in the main paper, except for the method-specific ones mentioned below. We refer to their papers for the settings of these specific hyperparameters. All additional results are listed in Table \ref{tab:5}.

For the Momentum Iterative Gradient-based Method (MIGM) \cite{dong2018boosting}, it integrates momentum into the iterative FGSM \cite{goodfellow6572explaining} and the update procedure of the adversarial patch $\delta$ can be formalized as follows,
\begin{equation}
\begin{split}
    \delta_{t+1} &= \delta_{t} + \alpha \cdot sign(g_{t+1}),\\
    g_{t+1} &= \mu \cdot g_t + \frac{\nabla_\delta J(\delta_t, y)}{\parallel \nabla_\delta J(\delta_t, y)\parallel_1}.
\end{split}
\label{eqn:2}    
\end{equation}
$J$ represents the loss function for the source crowd counting model. We set $\mu=1.0$ following \cite{dong2018boosting}.

For the Nesterov Iterative Gradient-based Method (NIGM) \cite{lin2019nesterov}, it utilizes Nesterov Accelerated Gradient \cite{nesterov1983method} to improve the attacking transferability. Following \cite{lin2019nesterov},  we can update our patch by slightly modifying the formula (\ref{eqn:2}) as follows,
\begin{equation}
\begin{split}
    \delta_{t+1} &= \delta_{t} + \alpha \cdot sign(g_{t+1}),\\
    g_{t+1} &= \mu \cdot g_t + \frac{\nabla_\delta J(\delta^{nes}_t, y)}{\parallel \nabla_\delta J(\delta^{nes}_t, y)\parallel_1},\\
    \delta^{nes}_t &= \delta_t + \alpha \cdot \mu \cdot g_t
\end{split}
\label{eqn:3}    
\end{equation}
We set $\mu=1.0$ and $\alpha=0.01$ referring to \cite{lin2019nesterov}.

Further, we combine the Translation-Invariant method \cite{lin2019nesterov} and NIGM, named TI-NIGM, which has much stronger transferability. Specifically, the accumulated gradients $g_{t+1}$ observe the following update rule,
\begin{equation}
\begin{split}
    g_{t+1} &= \mu \cdot g_t + \frac{W \ast \nabla_\delta J(\delta^{nes}_t, y)}{\parallel W \ast \nabla_\delta J(\delta^{nes}_t, y)\parallel_1},
\end{split}
\label{eqn:4}    
\end{equation}
where $W$ is the pre-defined gaussian kernel.

In addition to the method directly manipulating the model output, we also compare with a feature-level transfer-based attack named Neuron Attribution-based Attack (NAA) \cite{zhang2022improving} which can be formulated into solving the following constrained minimization problem,
\begin{equation}
\begin{split}
    \underset{\delta}{\min}\ f_\gamma((l - l') \cdot IA(l)),
\end{split}
\label{eqn:5}    
\end{equation}
where $l$ and $l'$ are the activation values of the neuron when the input is an adversarial image and a black image, respectively. $IA$ reflects Integrated Attention proposed in \cite{zhang2022improving}. $f_\gamma$ is a transformation function with hyperparameter $\gamma$ for distinguishing between positive and negative neuron attributions. Following \cite{zhang2022improving}, we set integrated step $n=30$ and $\gamma=1.0$. We choose MCNN-\emph{branch1,2,3}-(9), CSRNet-\emph{frontend}-(22), CAN-\emph{frontend}-(22), BL-\emph{features}-(35), DM-Count-\emph{features}-(35), SASNet-\emph{features5}-(9) as target layers to obtain $l$ and $l'$.

In addition to attacking with a single source model, we also consider the ensemble-based method. Referring to \cite{liu2016delving,tramer2017ensemble,dong2018boosting}, we first conduct an ensemble-based attack by Averaging Density (Avg-Dens). Specifically, for a target model, we generate adversarial patches using the other five models by taking the average predicted count as the loss as follows,
\begin{equation}
\mathcal L = \frac{1}{5}\sum_{k=0}^{5}\sum_{i,j}{{f_{\Theta}^k}_{i,j}(\mathbf{x}_{adv})}.
\label{eqn:6}
\end{equation}

Besides, we compare with another ensemble-based method called Meta Gradient Adversarial Attack (MGAA) \cite{yuan2021meta}. Specifically, we randomly sample four models from a source model zoo to compose different meta tasks and iteratively simulate a transfer-based black-box attack in each task. We set the number of iterations $K=5$ and the number of ensemble models $n=3$ in the meta-train step. To 
keep it comparable, we set the number of sample tasks $T=25$ and meta-test step $\beta = 0.01$.

\section{More results for the ablation study related to the loss functions}
\begin{figure*}[!ht]
\vspace{-0.1in}
\subfigbottomskip=-1.5ex
\subfigure[Results based on the source model MCNN]{
\includegraphics[width=0.95\linewidth]{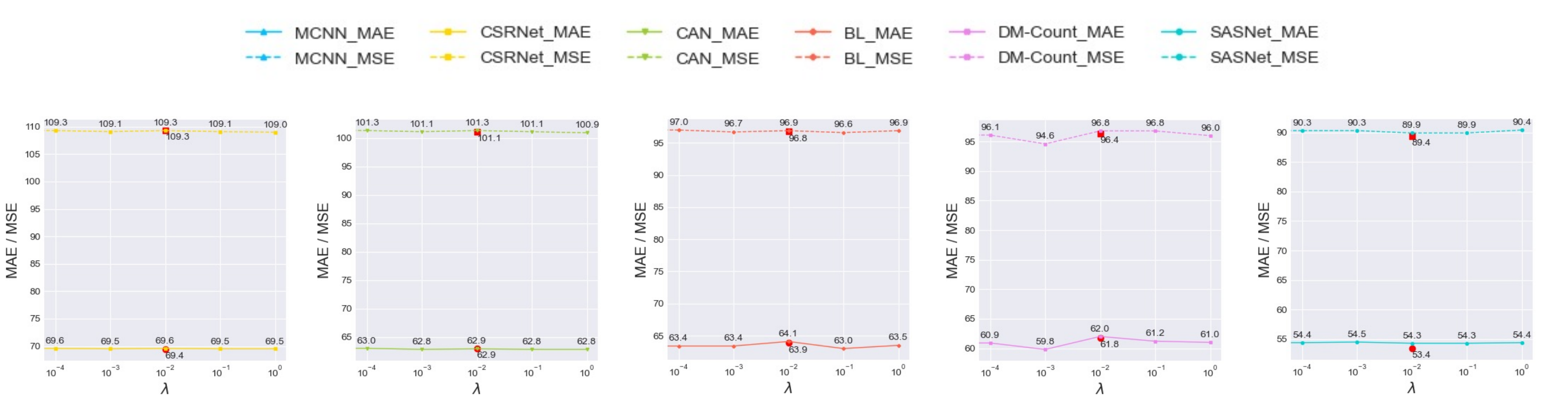}
}
\subfigure[Results based on the source model CAN]{
\includegraphics[width=0.95\linewidth]{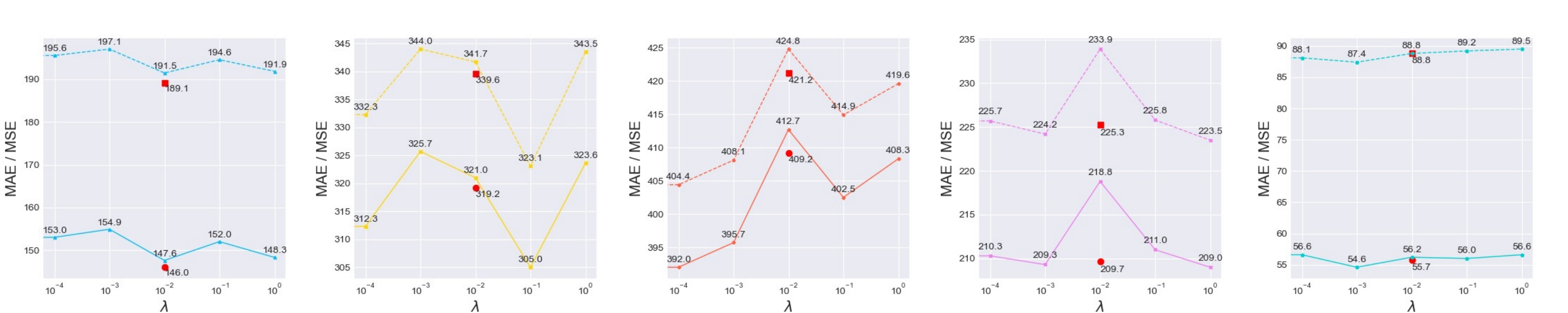}
}
\subfigure[Results based on the source model BL]{
\includegraphics[width=0.95\linewidth]{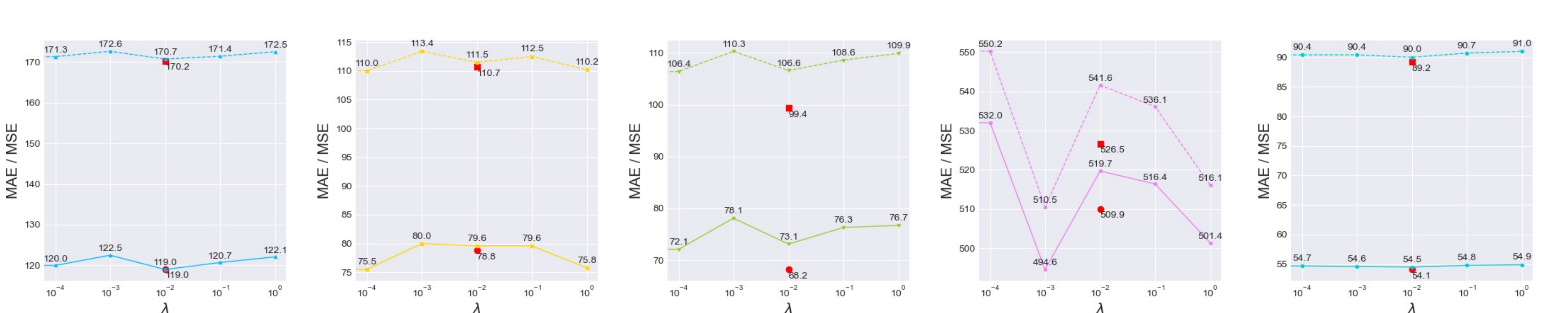}
}
\subfigure[Results based on the source model DM-Count]{
\includegraphics[width=0.95\linewidth]{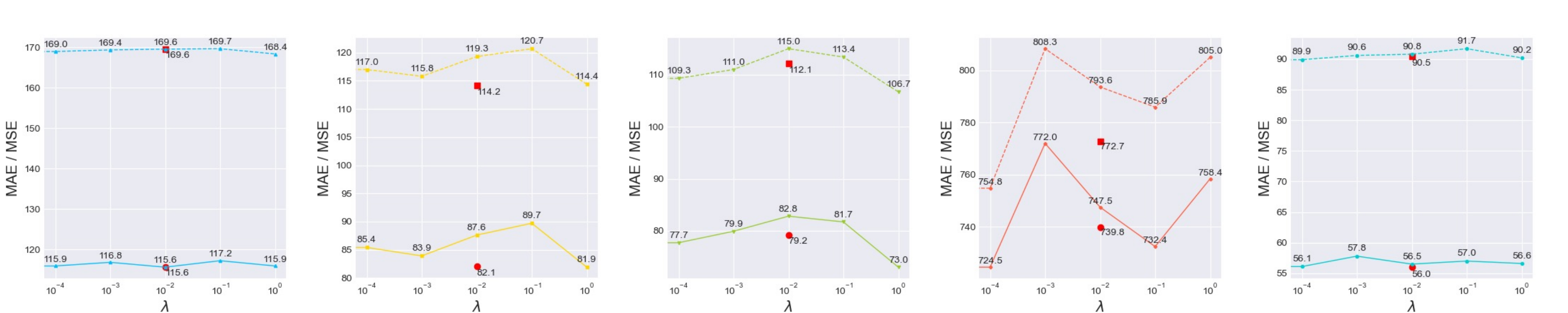}
}
\subfigure[Results based on the source model SASNet]{
\includegraphics[width=0.95\linewidth]{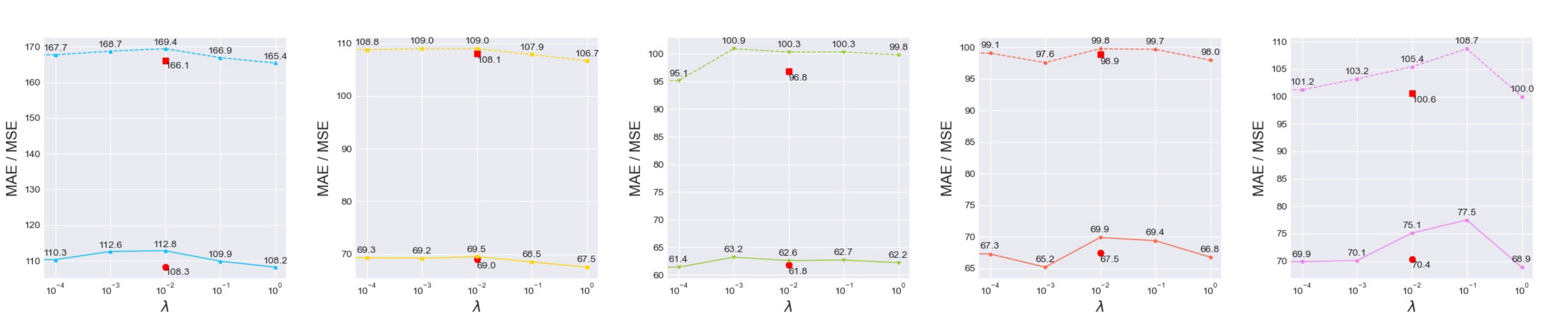}
}

\caption{The ablation study on the influence of $\lambda$. The ``red circle" and ``red square" means the MAE and MSE when $\lambda=0$. We show the MAE/MSE for the black-box attack on Shanghai Tech Part A. Higher MAE and MSE values indicate a stronger attack.}
\label{fig:ablation-supp}
\end{figure*}

We report the results of CSRNet \cite{li2018csrnet} in our main paper. In this section, we provide additional model results for the ablation study on two perception loss functions and the loss weight $\lambda$. Table \ref{tab:ablation-supp} and Figure \ref{fig:ablation-supp} respectively list the results for MCNN \cite{zhang2016single}, CAN \cite{liu2019context}, BL \cite{ma2019bayesian}, DM-Count \cite{wang2020distribution}, and SASNet \cite{song2021choose}. All results demonstrate the conclusions in the main paper.

\begin{table}[!ht]

\caption{Cross-dataset evaluation (results are shown as ``training dataset$\rightarrow$test dataset''). Adversarial training with once adversarial patch generation (OAT) will lead to better generalization than an iterative generation (IAT).}
\begin{center}
\setlength{\tabcolsep}{6mm}
\scriptsize
\begin{tabular}{ccc}
\toprule
{\centering\textbf{MAE / MSE}}&\multicolumn{2}{c}{\centering\textbf{Cross-dataset Evaluation}}\\
\midrule
{\centering\textbf{Method}}&{\centering\textbf{Part A$\rightarrow$Part B}}&{\centering\textbf{Part B$\rightarrow$Part A}}\\
\midrule
{Vanilla}
&22.8 / 34.3&142.4 / 241.3\\
{IAT}
&23.5 / 35.1&140.0 / 245.1\\
\textbf{OAT(Ours)}
&\textbf{17.5} / \textbf{27.5}&\textbf{129.8} / \textbf{220.5}\\
\bottomrule
\end{tabular}
\end{center}
\label{tab:at}
\end{table}

\begin{table}[!ht]
\vspace{-0.1in}
\caption{Robustness evaluation towards complex backgrounds. Adversarial training with once adversarial patch generation (OAT) will lead to better robustness than an iterative generation (IAT).}
\begin{center}
\setlength{\tabcolsep}{2mm}
\scriptsize
\begin{tabular}{cccc}
\toprule
{\centering\textbf{MAE / MSE}}&\multicolumn{3}{c}{\centering\textbf{Complex Backgrounds}}\\
\midrule
{\centering\textbf{Method}}&{\centering\textbf{Distractors}}&{\centering\textbf{Special Weathers}}&{\centering\textbf{Negative Samples}}\\
\midrule
{Vanilla}
&42.3 / 63.9&239.2 / 849.5&10.2 / 22.8\\
{IAT}
&40.4 / 64.2&242.2 / 840.7&{ 7.8} / 21.0\\
\textbf{OAT(Ours)}
&\textbf{38.4} / \textbf{61.1}&\textbf{228.9} / \textbf{833.1}&\textbf{ 7.5} / \textbf{18.3}\\
\bottomrule
\end{tabular}
\end{center}
\vspace{-0.1in}
\label{tab:complex}
\end{table}

\begin{table}[!ht]
\caption{MAE/MSE in cross-dataset evaluation. Lower MAE and MSE values indicate better generalization.}
\begin{center}
\setlength{\tabcolsep}{3mm}
\scriptsize
\subtable[Results of the MCNN models trained on Shanghai Tech Part A]{
\begin{tabular}{cccc}
\toprule
\centering\textbf{Method}&{\textbf{Shanghai Tech Part B}}&{\textbf{UCF-CC-50}}&{\textbf{Crowd Surveillance}}\\
\midrule
{Vanilla}
&50.1 / 62.8&441.8 / 700.9&159.0 / 210.1\\
\textbf{Ours}
&\textbf{33.1} / \textbf{48.8}&\textbf{431.4} / \textbf{669.6}&\textbf{87.1} / \textbf{148.0}\\
\bottomrule
\end{tabular}}

\subtable[Results of the MCNN models trained on Shanghai Tech Part B]{
\begin{tabular}{cccc}
\toprule
\centering\textbf{Method}&{\textbf{Shanghai Tech Part A}}&{\textbf{UCF-CC-50}}&{\textbf{Crowd Surveillance}}\\
\midrule
{Vanilla}
&178.7 / 265.9&624.0 / 950.9&167.1 / 221.9\\
\textbf{Ours}
&\textbf{157.8} / \textbf{265.0}&\textbf{503.7} / \textbf{724.8}& \textbf{33.3} / \textbf{55.5}\\
\bottomrule
\end{tabular}}
\label{tab:gen-mcnn}
\end{center}
\end{table}

\begin{table}[!ht]

\caption{MAE/MSE in robustness evaluation towards complex backgrounds.}
\begin{center}
\setlength{\tabcolsep}{3mm}
\scriptsize
\begin{tabular}{cccc}
\toprule
{\centering\textbf{Method}}&{\centering\textbf{Distractors}}&{\centering\textbf{Special Weathers}}&{\centering\textbf{Negative Samples}}\\
\midrule
{Vanilla}
&151.6 / 199.7&301.0 / 895.8&280.1 / 481.7\\
\textbf{Ours}
&\textbf{122.1} / \textbf{165.8}&\textbf{283.9} / \textbf{875.1}&\textbf{167.3} / \textbf{330.4}\\
\bottomrule
\end{tabular}
\end{center}
\label{tab:rob-mcnn}
\end{table}

\begin{table*}[!ht]
\caption{The ablation study on two perception loss functions. We show the MAE/MSE for the black-box attack on Shanghai Tech Part A. Higher MAE and MSE values indicate a stronger attack.}
\setlength{\tabcolsep}{9mm}
\scriptsize
\subtable[Results based on the source model MCNN]{
\begin{tabular}{cccccc}
\toprule
{\textbf{Loss}}&{\textbf{CSRNet}}&{\textbf{CAN}}&{\textbf{BL}}&{\textbf{DM-Count}}&{\textbf{SASNet}}\\
\midrule
{None}
&67.0 / 105.2
&59.9 / 94.1
&61.8 / 94.1	
&58.2 / 93.2
&52.8 / 86.2\\
{$\mathcal L_s$ w/o $W$}
&69.3 / 109.0
&62.8 / 101.1
&63.2 / 96.7
&60.2 / 94.4
&53.4 / 89.3\\
{$\mathcal L_s$}
&69.4 / 109.3
&62.9 / 101.1
&63.9 / 96.8
&61.8 / 96.4
&53.4 / 89.4\\
{$\mathcal L_p$}
&69.5 / 109.0
&62.9 / 101.2
&64.0 / 96.0
&60.8 / 95.5
&53.4 / 89.1\\
{\textbf{$\mathcal L_s + \lambda\mathcal L_p$}}
&\textbf{69.6} / \textbf{109.3}
&\textbf{62.9} / \textbf{101.3}
&\textbf{64.1} / \textbf{96.9}
&\textbf{62.0} / \textbf{96.8}
&\textbf{54.3} / \textbf{89.9}\\
\bottomrule
\end{tabular}}

\subtable[Results based on the source model CAN]{
\begin{tabular}{cccccc}
\toprule
{\textbf{Loss}}&{\textbf{MCNN}}&{\textbf{CSRNet}}&{\textbf{BL}}&{\textbf{DM-Count}}&{\textbf{SASNet}}\\
\midrule
{None}
&108.0 / 165.0
&67.0 / 105.2
&61.8 / 94.1	
&58.2 / 93.2
&52.8 / 86.2\\
{$\mathcal L_s$ w/o $W$}
&141.2 / 186.5
&303.6 / 317.7
&309.4 / 322.7
&173.6 / 191.3
&55.4 / 88.5\\
{$\mathcal L_s$}
&146.0 / 189.1
&319.2 / 339.6
&409.2 / 421.2
&209.7 / 225.3
&55.7 / 88.8\\
{$\mathcal L_p$}
&146.3 / 189.6
&300.0 / 320.1
&366.0 / 379.5
&210.6 / 225.7
&54.8 / 88.5\\
{\textbf{$\mathcal L_s + \lambda\mathcal L_p$}}
&\textbf{147.6} / \textbf{191.5}
&\textbf{321.0} / \textbf{341.7}
&\textbf{412.7} / \textbf{424.8}
&\textbf{218.8} / \textbf{233.9}
&\textbf{56.2} / \textbf{88.8}\\
\bottomrule
\end{tabular}}

\subtable[Results based on the source model BL]{
\begin{tabular}{cccccc}
\toprule
{\textbf{Loss}}&{\textbf{MCNN}}&{\textbf{CSRNet}}&{\textbf{CAN}}&{\textbf{DM-Count}}&{\textbf{SASNet}}\\
\midrule
{None}
&108.0 / 165.0
&67.0 / 105.2
&59.9 / 94.1	
&58.2 / 93.2
&52.8 / 86.2\\
{$\mathcal L_s$ w/o $W$}
&117.9 / 169.9
&76.8 / 109.7
&65.3 / 99.3
&492.3 / 509.9
&53.0 / 88.5\\
{$\mathcal L_s$}
&119.0 / 170.2
&78.8 / 110.7
&68.2 / 99.4
&509.9 / 526.5
&54.1 / 89.2\\
{$\mathcal L_p$}
&118.3 / 169.4
&76.9 / 110.8
&69.4 / 104.3
&505.7 / 525.1
&53.8 / 89.2\\
{\textbf{$\mathcal L_s + \lambda\mathcal L_p$}}
&\textbf{119.0} / \textbf{170.7}
&\textbf{79.6} / \textbf{111.5}
&\textbf{73.1} / \textbf{106.6}
&\textbf{519.7} / \textbf{541.6}
&\textbf{54.5} / \textbf{90.0}\\
\bottomrule
\end{tabular}}

\subtable[Results based on the source model DM-Count]{
\begin{tabular}{cccccc}
\toprule
{\textbf{Loss}}&{\textbf{MCNN}}&{\textbf{CSRNet}}&{\textbf{CAN}}&{\textbf{BL}}&{\textbf{SASNet}}\\
\midrule
{None}
&108.0 / 165.0
&67.0 / 105.2
&59.9 / 94.1	
&61.8 / 94.1
&52.8 / 86.2\\
{$\mathcal L_s$ w/o $W$}
&113.9 / 167.0
&77.5 / 111.1
&78.2 / 109.3
&726.7 / 773.6
&55.7 / 89.3\\
{$\mathcal L_s$}
&115.6 / 169.6
&82.1 / 114.2
&79.2 / 112.1
&739.8 / 772.7
&56.0 / 90.5\\
{$\mathcal L_p$}
&115.2 / 169.1
&81.4 / 111.4
&81.6 / 112.4
&728.6 / 768.5
&56.0 / 89.9\\
{\textbf{$\mathcal L_s + \lambda\mathcal L_p$}}
&\textbf{115.6} / \textbf{169.6}
&\textbf{87.6} / \textbf{119.3}
&\textbf{82.8} / \textbf{115.0}
&\textbf{747.5} / \textbf{793.6}
&\textbf{56.5} / \textbf{90.8}\\
\bottomrule
\end{tabular}}

\subtable[Results based on the source model SASNet]{
\begin{tabular}{cccccc}
\toprule
{\textbf{Loss}}&{\textbf{MCNN}}&{\textbf{CSRNet}}&{\textbf{CAN}}&{\textbf{BL}}&{\textbf{DM-Count}}\\
\midrule
{None}
&108.0 / 165.0
&67.0 / 105.2
&59.9 / 94.1	
&61.8 / 94.1
&58.2 / 93.2\\
{$\mathcal L_s$ w/o $W$}
&108.3 / 166.1
&68.9 / 106.6
&60.9 / 95.2
&65.3 / 98.2
&65.5 / 96.6\\
{$\mathcal L_s$}
&110.0 / 167.5
&69.0 / 108.1
&61.8 / 96.8
&67.5 / 98.9
&70.4 / 100.6\\
{$\mathcal L_p$}
&109.1 / 167.1
&67.2 / 105.5
&60.2 / 94.9
&62.7 / 96.3
&68.2 / 99.1\\
{\textbf{$\mathcal L_s + \lambda\mathcal L_p$}}
&\textbf{112.8} / \textbf{169.4}
&\textbf{69.5} / \textbf{109.0}
&\textbf{62.6} / \textbf{100.3}
&\textbf{69.9} / \textbf{99.8}
&\textbf{75.1} / \textbf{105.4}\\
\bottomrule
\end{tabular}}

\label{tab:ablation-supp}
\end{table*}

\section{Discussion for Different Adversarial Training Schemes}\label{scheme}
In this section, we plan to discuss the influence of different implementations of adversarial training. Specifically, we train two DM-Count models with our patches in two different adversarial training schemes, \ie, generating all adversarial examples based on the pre-trained model at the beginning or conducting the min-max optimization iteratively. Then, we evaluate their performance for generalization across datasets and robustness towards complex backgrounds.

As shown in Table \ref{tab:at} and Table \ref{tab:complex}, generating adversarial patches once at the beginning will achieve better performance. We conjecture that our patches may fail to capture satisfactory model-invariant features during the iteratively min-max optimization. Besides, iteratively generating adversarial examples and training the model will take obviously more time (622.1h) than once preparing (33.6h). Thus, we take an adversarial training scheme with once adversarial patch generation in our framework.

\section{More results for the model improvement}
In the main paper, we demonstrate the effectiveness of adversarial training with our PAP for the single-column method DM-Count \cite{wang2020distribution}. In this section, we use another crowd counting model MCNN \cite{zhang2016single}, a multi-column method, to evaluate our approach. Specifically, we conduct experiments to test the model generalization ability across datasets and robustness on scenes with complex backgrounds. Except for the crowd counting model, we all follow the same settings in the main paper. As shown in Table \ref{tab:gen-mcnn} and Table \ref{tab:rob-mcnn}, our method can generally benefit the model performance.

\end{document}